%% file: main.tex
\documentclass[12pt, a4paper]{article}

\usepackage[english]{babel} %
\usepackage[utf8]{inputenc}
\usepackage[T1]{fontenc}

\usepackage[left=2.5cm, right=2.5cm, top=3cm, bottom=3cm]{geometry} %
\usepackage{setspace} %
\usepackage{fancyhdr} %
\usepackage{float} %
\usepackage[font=footnotesize,labelfont=bf]{caption} %
\usepackage{rotating} %
\usepackage{placeins} %
\usepackage{multicol} %
\usepackage{color} %
\usepackage[
    left = \flqq,%
    right = \frqq,%
    leftsub = \flqq,%
    rightsub = \frqq%
]{dirtytalk}

\usepackage{endnotes}
\usepackage{graphicx} 
\graphicspath{ {./figures/} }
\usepackage{subcaption} %
\usepackage{tikz}
\usepackage[shell]{dot2texi}
\usetikzlibrary{shapes,arrows}
\usepackage{environ}

\usepackage{amsmath}
\usepackage{amssymb}
\usepackage{textcomp} %
\usepackage{relsize} %

\usepackage{mathptmx}
\DeclareMathAlphabet{\mathcal}{OMS}{cmsy}{m}{n} %
\usepackage[scaled=.90]{helvet}
\usepackage{courier}
\usepackage{csquotes}
\usepackage{lipsum}

\usepackage{booktabs} %
\usepackage{multirow}
\usepackage{makecell}
\usepackage{longtable}
\usepackage{lscape}
\usepackage{tabu}
\usepackage{booktabs}
\usepackage{longtable}
\usepackage{array}
\usepackage{multirow}
\usepackage{wrapfig}
\usepackage{float}
\usepackage{colortbl}
\usepackage{pdflscape}
 \usepackage{tabu}
\usepackage{threeparttable} 
\usepackage{threeparttablex}
 \usepackage[normalem]{ulem}
\usepackage{makecell}
\usepackage{xcolor}
\usepackage{rotating}

\usepackage[final]{pdfpages}

\usepackage[hyphens]{url}
\usepackage[hidelinks]{hyperref} %
\usepackage[all]{hypcap}
\usepackage{nameref}
\usepackage[backend=bibtex, natbib = true, style=chicago-authordate]{biblatex}
\bibliography{library.bib}

\title{
	{\textbf{ Learnability of Timescale Graphical Event Models }}\\[0.2cm]
	{\large \textbf{}}\\[4cm]
	{\large Technical Report on the PILGRIM Library and its Application on Timescale Graphical Event Models}\\[4cm]
	{\large Author: Philipp Behrendt}\\[0.5cm]
	{\large  Tutor: Prof. Dr. Philippe Leray}\\[2.3cm]
	{\large Nantes, 25\textsuperscript{th} May 2020}\\
}
\author{}
\date{}

\begin{document}

\maketitle
\thispagestyle{empty}

\newpage

\pagenumbering{Roman}

\tableofcontents
\newpage

\listoftables
\listoffigures

\newpage
\abstract{This technical report tries to fill a gap in current literature on Timescale Graphical Event Models. I propose and evaluate different heuristics to determine hyper-parameters during the structure learning algorithm and refine an existing distance measure. A comprehensive benchmark on synthetic data will be conducted allowing conclusions about the applicability of the different heuristics. 
}

\onehalfspacing

\pagenumbering{arabic}

\lhead{\bfseries\leftmark}
\rhead{}

\input{chapters/02_GEM.tex}

\input{chapters/03_methods.tex}

\input{chapters/04_analysis.tex}

\input{chapters/05_discussion.tex}

\FloatBarrier

\newpage

\lhead{\bfseries REFERENCES}
\rhead{}

\addcontentsline{toc}{section}{References}

\setcounter{biburllcpenalty}{7000}
\setcounter{biburlucpenalty}{8000}
\setcounter{biburlnumpenalty}{8000}

\printbibliography

\FloatBarrier

\newpage

\lhead{\bfseries APPENDIX A: SYNTHETIC DATA}
\rhead{}

\addcontentsline{toc}{section}{Appendix A: Synthetic Data}
\input{chapters/appendix_a.tex}

\end{document}

%% file: chapters/02_GEM.tex
\section{Graphical Event Models}

This chapter introduces the class Graphical Event Models and in particular Timescale Graphical Event Models \citep{Gunawardana2016}. After a reminder of the general framework and its notation, I will recap the structure learning algorithm and discuss different heuristics to choose hyper-parameters. Further, I briefly explain how to generate synthetic data. Finally, I propose a refined distance measure to evaluate how similar two Timescale Graphical Event Models are. For the sake of consistency, definitions and notations are adopted from the original work of \citet{Gunawardana2016}.

 Event streams and their temporal dynamics can be represented as a \textit{multivariate temporal point process} and the literature offers several advanced methods such as Continuous Time Bayesian Networks \citep{ctbn2002}, Poisson Networks \citep{pnetwork2005}, Conjoint Piecewise-Constant Conditional Intensity Models \citep{Parikh2012}, or Multiplicative-Forest Point Processes \citep{Weiss2013}. They commonly share the concept of \textit{conditional intensity functions} to express the rate at which a specific event occurs, conditioned on previous event occurrences.

Graphical Event Models (GEMs) \citep{Didelez2008, Meek2014,Gunawardana2016} provided a framework that generalizes before-mentioned models. Moreover, \citet{Gunawardana2016} showed that GEMS can universally approximate any smooth multivariate temporal point process. GEMs provide a compact graphical representation of such process where different events are represented as nodes and an edge from node $A$ to node $B$ implies that the appearance of event $A$ has some influence on the occurrence of event $B$. In addition to this qualitative information about temporal dependencies, GEMs also contain quantitative information about these dynamics in terms of conditional intensity functions.

\subsubsection*{Preliminaries}

\citet{Gunawardana2016} denote a stream of events as $(t, l) \in \mathbb{R}^+ \times \mathcal{L}$, each of which has a timestamp $t > 0$ and a label $l$ taken from a finite label vocabulary $\mathcal{L}$. This yields a sequence  
$\{(t_{1}, l_{1}), \dots, (t_{i}, l_{i}), \dots, (t_{n}, l_{n})\}$,  where $t_{0} = 0 < t_{i} < t_{i+1} < t^{*}$ and $1 \leq i \leq n-1$. Let further be $x_{t^{*}}$ the sequence of events $\{(t_i, l_i): t_i < t^*\}$ until time $t^*$ and $h_i$ the $i$th history $h_i = (t_1, l_1), ... , (t_{i-1}, l_{i-1})$. 

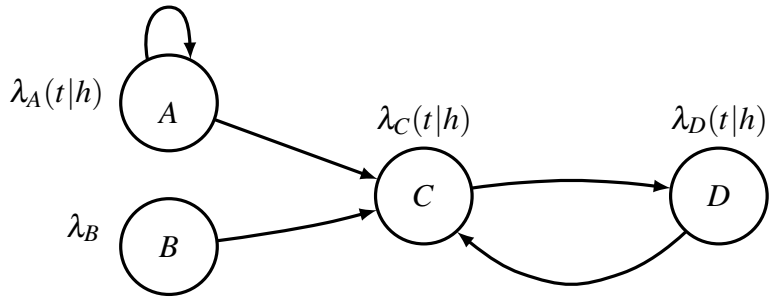
\begin{figure}[htb]
\centering
\input{figures/GEM.tex}

\caption[Illustration of a Graphical Event Model]{Example of a Graphical Event Model with $4$ distinct nodes and the corresponding conditional intensity functions}
\label{GEM-Example}
\end{figure}

Then, a \textit{Graphical Event Model} is defined as a directed graph $\mathcal{G} = (\mathcal{L},E)$. Its likelihood for a given event stream $x_{t^*}$ can be written as

\begin{equation}
p(x_{t^*} | t^*) = \prod_{i=1}^{| x_{t^*} |} \lambda_{l_i}(t_i | h_i) \prod_{i=1}^{| x_{t^*}| +1} e^{-\sum_{l \in \mathcal{L}} \int_{t_i - 1}^{t_i} \lambda_l(\tau | h_i) d\tau}
\label{eqGEM}
\end{equation}

with  $\lambda_l(t|h) > 0 $ as the \textit{conditional intensity function} of event $l$ at time $t$ given the history $h$. It defines the rate of event $l$ to occur at time $t$ depending on the observed history $h$. A multivariate temporal point process is Markovian with respect to a GEM if 

\begin{equation}
\lambda_l(t | h) = \lambda_l(t | [h]_{Pa(l)})
\end{equation}
where $Pa(l)$ are the parents of $l$ in $\mathcal{G}$. It states that the occurrence of event $l$ only depends on its parents in $\mathcal{G}$. Figure \ref{GEM-Example} provides a simple example of a GEM with 4 labels $A,B,C,$ and $D$. One can easily read off the dependency from the graph. For instance, the rate $\lambda_A(t | h)$ for event $A$ only depends of its own history. Event $C$ however, has three parents and its rate  $\lambda_C(t | h)$ depends on the previous occurrences of $A,B,$ and $D$. In contrast, event $B$ has no parents, i.e., it does not depend of any event in the history. Thus, the rate   $\lambda_B(t | h)$ is constant and  $B$ forms a homogeneous Poisson process. Accordingly, the rate $\lambda_D(t | h)$ solely depends of the history of event $C$. 

\subsection{Timescale Graphical Event Model}
\label{ch:TGEM}
Additionally to their contribution to the GEM framework, \citet{Gunawardana2016} proposed  \textit{Timescale Graphical Event Models} (TGEMs), a specific case of GEM where 
the temporal range and granularity of each dependency is explicitly stated. Accordingly, each edge $e \in E$ is enriched with additional information to which \citet{Gunawardana2016} refer as \textit{timescale}. A timescale is defined as a set $T$ of half-open intervals $(a,b]$ (with $a \geq 0$ and $b>a$) that form a partition of some interval $(0,t_h]$, where $t_h$ is the highest value of $T$ and denoted as \textit{horizon}. 

Consequently, a TGEM is defined as 
$M = (\mathcal{G}, \mathcal{T})$ consisting of a GEM $\mathcal{G} = (\mathcal{L},E)$ and a set of timescales $\mathcal{T} = {T_e}_{({e \in E})}$ corresponding to the edges $E$ of the graph $\mathcal{G}$. The conditional intensity functions are given by 
\begin{equation}
\lambda_l(t | h) = \lambda_{l,c_l(h,t)}
\end{equation}
where the index $c_l(h,t)$ is the \textit{parent count vector} of $l$. It contains the number of occurrences of the parents with respect to the corresponding timescales. $C_l$ denotes the set of all possible parent count vectors of label $l$. Like \citet[p.568]{Gunawardana2016}, I assume throughout this work that all \textit{parent count vectors} are bounded by $1$, making them binary. Hence, it is only of importance whether a parent has occurred or not within the respective interval on the timescale.

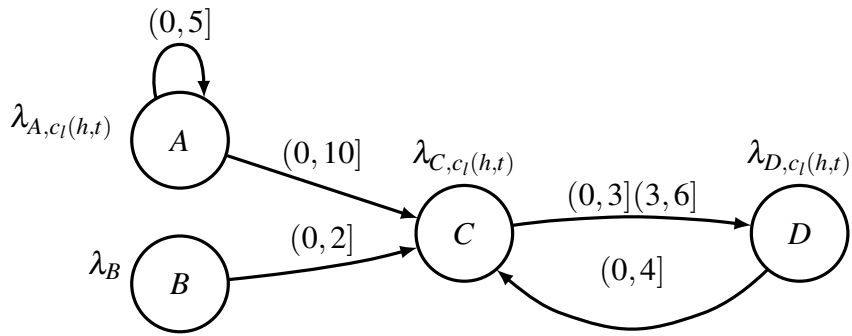
\begin{figure}[htb]
\centering

\input{figures/TGEM.tex}

\caption[Illustration of a Timescale Graphical Event Model]{Example of a Timescale Graphical Event Model with $4$ distinct nodes and the corresponding conditional intensity functions}
\label{TGEM-Example}
\end{figure}

Consider the TGEM in Figure \ref{TGEM-Example} which extends the example GEM from above with an arbitrary set of timescales $T$. One can denote the set of parent count vectors $C_l$ for each node:  $C_A = \{ 0,1\}$, $C_B =  \emptyset$,  $C_C = \{ 0,1\}^3$, $C_D = \{ 0,1\}^2$. For instance, the parent count vector $c_D(h,t)$ indicates whether event $C$ happened during the intervals $[t-3,t)$ and $[t-6,t-3)$. Conversely, $c_C(h,t) = [0,1,1]$ means that event $A$ did not occur during $[t-10,t)$, but event $B$ occurred during $[t-2,t)$ and event $D$ occurred during $[t-4,t)$. Following this notation, one can easily list the different conditional intensities for each event: for event $D$, these would be    
$\lambda_{D,00}$ , $\lambda_{D,01}$ , $\lambda_{D,10}$ , and  $\lambda_{D,11}$. The same logic applies to the other nodes in Figure \ref{TGEM-Example}, except $B$ which has no parents and its conditional intensity is therefore simplified to $\lambda_{B}$.

\citet{Gunawardana2016} assume that a conditional intensity $\lambda_{l,c_l(h,t)}$ is  constant, thus making the conditional intensity functions piecewise-constant. Finally, the likelihood of a TGEM $M$ for a given event stream $x_{t^*}$ can be expressed as

\begin{equation}
p(x_{t^*} | t^*) = \prod_{l \in \mathcal{L}}\prod_{j \in C_l} \lambda_{l,j}^{n_{t^*,l,j}(x_{t^*})}e^{-\lambda_{l,j}d_{t^*l,j}(x_{t^*})}
\label{Likelihhod_TGEM}
\end{equation}

where  $n_{t^*,l,j}(x_{t^*})$ is the number of occurrences of event $l$ within the parent configuration\footnote{As the parent count vector encodes a specific setting of parents for a given node, I will use the term parent configuration as a more intuitive denomination} $j$, and $d_{t^*l,j}(x_{t^*})$ is the duration of this parent configuration $j$. Since at each time $t$, exactly one parent configuration is active for a given node $l$, $d_{t^*l,j}(x_{t^*})$ builds a partition of $t^*$ and therefore $\sum_{{j \in C_l}}d_{t^*l,j}(x_{t^*}) = t^*$. Equivalently,  $\sum_{{j \in C_l}}n_{t^*,l,j}(x_{t^*}) = n_{t^*,l}(x_{t^*})$.

\subsection{Structure Learning of TGEMs}

To learn the structure and parameter of a TGEM  $M$ from some data $x_{t^*}$, \citet{Gunawardana2016} proposed an asymptotically consistent greedy algorithm that follows a score-based search approach. Its core idea is to define a model criterion that evaluates how well $M$ fits $x_{t^*}$, and to traverse the space of TGEMs by iteratively checking whether modifying the graph with an elementary operator would improve the score. 
The proposed score adapts from the \textit{Bayesian Information Criterion} (BIC) \citep{schwarz_bic} and is defined as follows:
 
\begin{equation}
BIC_{t^*}(\mathcal{M}) = log(p(x_{t^*} | t^* ; M, \widehat{\lambda}_{t^*,l,j}(x_{t^*}))) - \sum_{l \in \mathcal{L}} | C_l | log(t^*)
    \label{BICscore}
\end{equation}

with 
\begin{equation}
\widehat{\lambda}_{t^*,l,j}(x_{t^*}) = \frac{n_{t^*,l,j}(x_{t^*})}{d_{t^*l,j}(x_{t^*})}
\label{ML_estimator}
\end{equation} 

as the maximum likelihood estimate (MLE) for each parent configuration. The score can be viewed as a combination of log-likelihood of $M$ given the data $x_{t^*}$ and a regularization term that penalizes the complexity of the model.

The proposed learning algorithm follows two steps: in the \emph{Forward search} edges are added and timescales are refined, whereas the \emph{Backward search} tries to simplify the model. \citet{Gunawardana2016} make use of a subfamily of TGEMs which they call Recursive TGEMs. It refers to any TGEM that can be build by performing \emph{recursively} elementary operators $\mathcal{O} = \{add, split, extend\}$, starting from an empty model. These elementary operators have the following definitions: 

\begin{itemize}

	\item  $O_{add}(e)$ adds a non-existing edge $e$ to $E$ with a timescale $\mathcal{T} = (0, h_{def}]$ where $h_{def}$ is a default horizon 
	
	\item $O_{split}(\mathcal{T}_{e})$ splits an interval $(a,b]$ of a  timescale  of an  existing edge and substitutes it with $(a,\frac{a+b}{2}], (\frac{a+b}{2},b]$

	\item $O_{extend}(e)$ extends the horizon of an existing edge by appending $(h,2h]$ to the timescale

\end{itemize}

The \emph{Forward search} starts from the empty model $\mathcal{M}_{0}$ and 
computes the neighborhood until convergence of BIC. This state is denoted as $\mathcal{M_{FS}}$. The neighborhood $\mathcal{N}_{FS}(\mathcal{M})$ of $\mathcal{M}$ is the set of RTGEMs that can be reached with one elementary operator. Formally, $\mathcal{M'} \in \mathcal{N}_{FS}(\mathcal{M}) \Leftrightarrow \exists O \in \mathcal{O}$ such as $O(\mathcal{M}) = \mathcal{M'}$ \citep{Monvoisin2019}. 

The \emph{Backward search} starts with $\mathcal{M_{FS}}$ and computes the neighborhood until convergence of BIC. The neighborhood $\mathcal{N}_{BS}(\mathcal{M})$ is the set of RTGEMs that can be reached with the inversion of one elementary operator. Formally, $O(\mathcal{M'}) = \mathcal{M}$.

\subsection{Choice of Default Horizon}
\label{Horizon Choice}

An essential aspect that \citet{Gunawardana2016} did not address in their work is the choice of the  default horizon for $O_{add}(e)$. As the Forward Search starts from an empty model $\mathcal{M}_0$, the initial neighborhood $\mathcal{N}_{FS}(\mathcal{M}_0)$ consists only of RTGEMs that are reached by $O_{add}(e)$, since there exist no edges yet to be extended or split. Thus, the choice of $h_{def}$ is critical and a too small or too large value could possibly inhibit the learning process. 

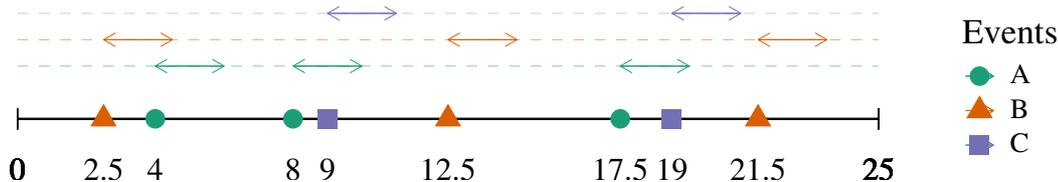
\begin{figure}[htb]
\centering
\input{figures/time_line.tex}
\caption[Illustration of an event stream]{Example of an event stream with three distinct events. The double-headed arrows indicate the hypothetical temporal ranges for the corresponding events with a default horizon $h_{def} = 2$.}
		\label{plot:example_timeline}
\end{figure}

The example of Figure \ref{plot:example_timeline} illustrates this problem. It depicts an event stream with three distinct events $A$, $B$, and $C$ until $t^* = 25$. Consider a global default horizon $h_{def} = 2$ for all edges. The double-headed arrows indicate the interval in which the occurrence of an event $l$  would have an impact w.r.t. $h_{def} = 2$. 

It is straight forward that such a global choice is inadequate for the given example\footnote{In fact, a global constant would be to some extent analog to a fixed lag in time-series analysis}. For instance, a dependency between $A \rightarrow C$ could be  detected as $C$ is preceded by $A$ within the interval $[t-2,t)$. On the contrary, a dependency from $C \rightarrow B$ would have never been found during the Forward Search as $B$ is never preceded by $C$ within the interval $[t-2,t)$, even though using a  
$h_{def} = 4$ would possibly find a dependency. 

The previous example illustrated why a global constant is an inappropriate choice for the default horizon. Alternatively, one might specify a default horizon for each edge individually $h_{def}(e)$. As the complexity of this increases quadratically in the number of different events $ |\mathcal{L}|$, an expert-knowledge based approach or a manual specification of each $h_{def}(e)$ is costly and infeasible for large graphs. Thus, a data-driven solution deems appropriate. 

A relevant measure for event streams are inter-event times. I adapt the notation of \citet{Bhattacharjya2018} who define the inter-event times $\{t_{ZX}\}$ from event $Z$ to $X$ as the set of times from the most recent occurrence of $Z$, if $Z$ occurred, to every occurrence of $X$. Further, let $\{t_{XX}\}$ the inter-event times between $X$ and the time from the last occurrence of $X$ to $t^*$. In Figure \ref{plot:example_timeline}, $\{t_{CB}\} = \{2.5, 3.5 \}$,$\{t_{AA}\} = \{4,7.5,   9.5\}$, and $\{t_{AC}\} = \{1, 1.5\}$. 
Based on this notion, I propose two heuristics to determine the default horizon.

\subsubsection*{Quantile Heuristic}
One naive but computational inexpensive approach would be to take a specific quantile $q$ of the ordered inter-event times between parent and child of the considered edge $e$ as the default horizon $h_{def}(e)$. I will refer to this approach as the \textit{quantile heuristic}.

Consider the example in Figure \ref{plot:example_timeline}, opting for the median $q = 0.5$, the default horizon for the edge from $A$ to $A$ would equal $7.5$\footnote{This is the median of $t_{AA}$}.
By choosing a low value for $q$ and one implicitly assumes that the effect of the parent event has a shorter duration and thus affected only few of the child event occurrences. Conversely, the higher $q$, the more child events are assumed to be affected.

\subsubsection*{Proximal Heuristic}
A more sophisticated approach is adapted from the work of  \citet{Bhattacharjya2018} on Proximal Graphical Event Models (PGEMs), a special kind of TGEMs allowing only one timescale per edge. The idea is to find a default horizon\footnote{\citet{Bhattacharjya2018} denote it as \textit{optimal window}} that maximizes the likelihood as given in equation \ref{Likelihhod_TGEM}. It is shown that this is equivalent to maximize the Kullback-Leibler-Divergence between the count-based probabilities 
$\dfrac{n_{t^*,l,j}(x_{t^*})}{n_{t^*,l}(x_{t^*})}$ and the duration-based probabilities $\dfrac{d_{t^*l,j}(x_{t^*})}{t^*}$ \citep{Bhattacharjya2018} . For simplicity, consider the likelihood for only one node $l$ of equation \ref{Likelihhod_TGEM}. Thus, its \textit{log}-likelihood after substituting $\lambda_{l,j}$ with equation \ref{ML_estimator} can be rearranged to

\begin{align}
LogL(x_{t^*},l | t^*) =  \sum_{j \in C_l} n_{t^*,l,j}(x_{t^*}) \ln \dfrac{n_{t^*,l,j}(x_{t^*})}{d_{t^*l,j}(x_{t^*})} - \sum_{j \in C_l} n_{t^*,l,j}(x_{t^*})
\label{KL-Divergence}
\end{align}
As the second term is constant (number of $l$-events), it does not affect the maximization. Expanding equation \ref{KL-Divergence} with the constants $ln( n_{t^*,l,j}(x_{t^*}))) ^{-1} $ and $ln(d_{t^*l,j}(x_{t^*}))$ yields the formula of the KL-Divergence. The intuition behind this approach is to find a default horizon where the distribution of event counts differs maximally from the corresponding duration across the parent configurations $j \in C_l$. 

Still, this remains an optimization problem with a non-linear objective-function. However, \citet{Bhattacharjya2018} proved that for a node $X$ with a single parent $Z$, the maximizing horizon belongs to or is a left limit of the candidate set $H^* = \{t_{ZX}\} \bigcup max\{t_{ZZ }\}$. This is due to the fact that the event counts only change at the inter-event times $ t_{ZX }$ and are further upper bounded by $max\{t_{ZZ }\}$. For a formal proof, please refer to the \citet{Bhattacharjya2018}. 

Hence, to determine the default horizon for edge $e$ from $Z$ to $X$, I exhaustively search over $H^*$ and choose the value that maximizes the KL-Divergence.  
I will refer to this approach as the \textit{proximal heuristic}.

\subsection{Sampling from a TGEM}
\label{Sampling}

The creation of synthetic data from a TGEM $M$ until time $t^{end}$ can be generalized from the approach of Poisson-Networks \citep{pnetwork2005}. 
For a node $l$ without any parents, the inter-arrival times are simply drawn from an exponential distribution with a constant $\lambda_l$. For nodes with parents, the conditional intensities $\lambda_{l,c_l}$ depend on the current parent configuration and their occurrences must be known. In this case, rejection sampling is used. An inter-arrival time $\tau_l$ is drawn from an exponential distribution with the current $\lambda_{l,c_l}$  and only accepted if it appears before time $\hat{t_l}$ denoting the next change of the node's parent configuration. Otherwise, the sampling time is updated to $\hat{t_l}$ and $\lambda_{l,c_l}$ to the new parent configuration. For cyclic structures, similar considerations apply, however, these nodes must be sampled simultaneously. Inter-arrival times $\tau_l $ for each involved node are sampled with their corresponding rates $\lambda_{l,c_l}$. All values except the minimum are rejected, as the $ \min(\tau_l)$ might have changed the rates of the other nodes. However, $ \min(\tau_l)$ is only accepted, if it is happens before  $min (\hat{t_l})$ denoting the first change of parent configuration for any node of the cyclic structure (as this might again have changed the rate for this node). Otherwise, the sampling time is updated to $min (\hat{t_l})$ and accordingly the  conditional intensities. 
As mutual dependencies require simultaneous sampling and parents must be sampled prior to their children, \citet{pnetwork2005} propose the following procedure to sample efficiently: First, retrieve the strongly connected components (SCC\footnote{A SCC is a directed sub-graph where there exists a path between every pair of nodes.}) of a TGEM. Secondly, let each component represent a node in a directed acyclic graph, from which the nodes/components will be sampled in topological order.

\subsection{Distance Measure between RTGEMs}
\label{Distance Measure}

\citet{Antakly19} proposed an extension of the usual Structural Hamming Distance (SHD)\footnote{SHD is commonly used to assess how much Graphical Models such as Bayesian networks differ in their structure (e.g., \citet{Tsamardinos2006})} as global measure for the distance between two RTGEMs. Its overall idea is to add $1$ to the global distance, if an edge exist in only one of the two graphs, and a value $d \in [0, 1)$  accounting for the difference between the timescales of edges that appear in both graphs. Thus, for $\mathcal{M}_{1} = ((\mathcal{L}, E_{1}), \mathcal{T}_{1})$ and $\mathcal{M}_{2} = ((\mathcal{L}, E_{2}), \mathcal{T}_{2})$ with the same set of labels, it is defined as

\begin{equation}
    d(\mathcal{M}_{1}, \mathcal{M}_{2}) = \sum_{e \in E_{sd}} 1 + \sum_{e \in E_{inter}} d_e(\mathcal{T}_{1,e}, \mathcal{T}_{2,e}),
\end{equation}

where $E_{sd} = E_1 \triangle E_2$ and $E_{inter} = E_1 \cap E_2$. Let  $\mathcal{T}_{i,e}$ be the timescales for edge $e$ in model $\mathcal{M}_{i}$ and $v_{i}$ the corresponding set of endpoints\footnote{Alternative way to represent timescales. $\mathcal{T} = (0,a], (a,b], (b,c]$ is equivalent to $v = [0, a, b, c]$.} of model $\mathcal{M}_{i}$. The \textit{elementary} distance between the timescales is defined by:

\begin{equation}
    d_e(\mathcal{T}_{1,e}, \mathcal{T}_{2,e}) = \frac{v_{nid}}{v_{nid} + v_{id}}
\end{equation}

with  $v_{nid} = | v_{1} \triangle v_{2}|$ and $v_{id} = | v_{1} \cap v_{2}|$ as number of endpoints that exist in only one or both timescales, respectively. 

However, this measure considers the timescales as sets and neglects its quantitative information. In particular, it is inadequate in cases where the default horizon and consequently, the timescales are determined in a data-driven way. Consider three timescales $ \mathcal{T}_{A,e}, \mathcal{T}_{B,e}, \mathcal{T}_{C,e}$ and their sets of endpoints $v_A = [0, 2, 4]$, $v_B = [0, 1.99, 3.98]$,  and $v_C = [0, 16, 32]$. Then both, $d_e(\mathcal{T}_{A,e}, \mathcal{T}_{B,e})$ and $d_e(\mathcal{T}_{A,e}, \mathcal{T}_{C,e})$ yield $0.8$, even though $v_A$ and $v_B$ cover approximately the same time intervals (whereas $v_C$ does not).

Thus, I propose a refinement for the \textit{elementary distance} that incorporates these quantitative aspects. The idea is to find matches (if existing) between the endpoints of the two timescales based on the mutual minimal absolute difference. Formally,  
$$\forall (i,j), v_{1_i} \in v_1, v_{2_j} \in v_2 \quad m =  \{ (v_{1_i},v_{2_j} ): cl(v_{1_i},v_2 ) = v_{2_j} \wedge  cl(v_{2_j},v_1 ) = v_{1_i}  \}$$ 
with $cl$ as function to find the closest element to $v_{1_i}$ in $v_2$:  $cl(v_{1_i},v_2) = argmin_{v_{2_j}}(| v_{1_i} - v_{2_j}|)$

For each pair $p$ in $m$, the sum of the relative differences (scaled by its minimum) is taken into account. For unmatched endpoints a value of $1$ is considered. Finally, the corresponding terms are scaled with the number matched endpoints $ e_m = | m |$ and the number of unmatched endpoints $e_{nm}$ , respectively. 

\begin{equation}
    d^*_e(\mathcal{T}_{1,e}, \mathcal{T}_{2,e}) =    \frac{e_m}{e_m + e_{nm} }\left(\sum_{p \neq (0,0) \in m} \frac{| v_{1_i} - v_{2_j} |}{ \min(v_{1_i},v_{2_j})} \right) +  \frac{e_{nm}}{e_m + e_{nm}}
\end{equation}

For the example from above, this refinement of the \textit{elementary distance} leads to  $d^*_e(\mathcal{T}_{A,e}, \mathcal{T}_{B,e}) = 0.003$ and $d^*_e(\mathcal{T}_{A,e}, \mathcal{T}_{C,e}) = 0.8$.  

\subsection{Implementation as C++ Library}

I actively contributed to a C++ Library (PILGRIM\footnote{ \url{http://pilgrim.univ-nantes.fr}, visited on 22/02/2020}) maintained by the Data User KnowledgE (DUKe) research group of the LS2N laboratory in Nantes, France.  I implemented various of the before-mentioned concepts, including the sampling, the different heuristics to determine the default horizon, the refined distance function. Further, I contributed several utilities such as a caching for the structure learning, parallel computation for horizon heuristics, a random TGEM generator, and plotting. The library is still under development.

%% file: figures/GEM.tex
\begin{tikzpicture}[>=latex,line join=bevel,transform shape]
  \pgfsetlinewidth{1bp}
\pgfsetcolor{black}
  \draw [->,very thick] (73.984bp,88.29bp) .. controls (72.051bp,98.389bp) and (74.723bp,108.0bp)  .. (82.0bp,108.0bp) .. controls (86.662bp,108.0bp) and (89.434bp,104.06bp)  .. (90.016bp,88.29bp);
  \definecolor{strokecol}{rgb}{0.0,0.0,0.0};
  \pgfsetstrokecolor{strokecol}
  \draw (82.0bp,115.5bp) node {$$};
  \draw [->,very thick] (98.984bp,65.743bp) .. controls (113.22bp,60.5bp) and (133.78bp,52.921bp)  .. (159.84bp,43.321bp);
  \draw (129.5bp,65.5bp) node {$$};
  \draw [->,very thick] (194.89bp,39.981bp) .. controls (200.66bp,40.808bp) and (207.08bp,41.591bp)  .. (213.0bp,42.0bp) .. controls (229.85bp,43.164bp) and (234.15bp,43.164bp)  .. (251.0bp,42.0bp) .. controls (253.59bp,41.821bp) and (256.27bp,41.571bp)  .. (269.11bp,39.981bp);
  \draw (232.0bp,49.5bp) node {$$};
  \draw [->,very thick] (99.916bp,20.166bp) .. controls (111.66bp,21.726bp) and (127.32bp,24.068bp)  .. (141.0bp,27.0bp) .. controls (143.85bp,27.612bp) and (146.82bp,28.318bp)  .. (159.53bp,31.694bp);
  \draw (129.5bp,34.5bp) node {$$};
  \draw [->,very thick] (274.68bp,23.55bp) .. controls (268.28bp,17.557bp) and (259.93bp,11.15bp)  .. (251.0bp,8.0bp) .. controls (235.07bp,2.3827bp) and (228.93bp,2.3827bp)  .. (213.0bp,8.0bp) .. controls (207.42bp,9.9686bp) and (202.06bp,13.21bp)  .. (189.32bp,23.55bp);
  \draw (232.0bp,26.8bp) node {$ $};
  \draw (232.0bp,11.8bp) node {$$};
\begin{scope}
  \definecolor{strokecol}{rgb}{0.0,0.0,0.0};
  \pgfsetstrokecolor{strokecol}
  \draw [very thick] (82.0bp,72.0bp) ellipse (18.0bp and 18.0bp);
  \draw (82.0bp,68.3bp) node {$A$};
  \draw (40.0bp,75.3bp) node {$ \lambda_A(t | h) $};
\end{scope}
\begin{scope}
  \definecolor{strokecol}{rgb}{0.0,0.0,0.0};
  \pgfsetstrokecolor{strokecol}
  \draw [very thick] (177.0bp,37.0bp) ellipse (18.0bp and 18.0bp);
  \draw (177.0bp,37.0bp) node {$C$};
  \draw (177.0bp,65.3bp) node {$ \lambda_C(t | h) $};

\end{scope}

\begin{scope}
  \definecolor{strokecol}{rgb}{0.0,0.0,0.0};
  \pgfsetstrokecolor{strokecol}
  \draw [very thick] (287.0bp,37.0bp) ellipse (18.0bp and 18.0bp);
  \draw (287.0bp,37.0bp) node {$D$};
  \draw (287.0bp,65.3bp) node {$ \lambda_D(t | h) $};
\end{scope}
\begin{scope}
  \definecolor{strokecol}{rgb}{0.0,0.0,0.0};
  \pgfsetstrokecolor{strokecol}
  \draw [very thick] (82.0bp,18.0bp) ellipse (18.0bp and 18.0bp);
  \draw (82.0bp,18.0bp) node {$B$};
  \draw (50.0bp,25.3bp) node {$ \lambda_B $};
\end{scope}
\end{tikzpicture}

%% file: figures/TGEM.tex
\begin{tikzpicture}[>=latex,line join=bevel,transform shape]
  \pgfsetlinewidth{1bp}
\pgfsetcolor{black}
  \draw [->,very thick] (9.0212bp,87.916bp) .. controls (6.679bp,98.15bp) and (9.6719bp,108.0bp)  .. (18.0bp,108.0bp) .. controls (23.465bp,108.0bp) and (26.633bp,103.76bp)  .. (26.979bp,87.916bp);
  \definecolor{strokecol}{rgb}{0.0,0.0,0.0};
  \pgfsetstrokecolor{strokecol}
  \draw (18.0bp,115.5bp) node {$(0,5]$};
  \draw [->,very thick] (35.504bp,66.22bp) .. controls (52.251bp,60.691bp) and (77.751bp,52.271bp)  .. (106.82bp,42.672bp);
  \draw (71.0bp,66.5bp) node {$(0,10]$};
  \draw [->,very thick] (141.89bp,39.981bp) .. controls (147.66bp,40.808bp) and (154.08bp,41.591bp)  .. (160.0bp,42.0bp) .. controls (183.5bp,43.624bp) and (189.5bp,43.624bp)  .. (213.0bp,42.0bp) .. controls (215.59bp,41.821bp) and (218.27bp,41.571bp)  .. (231.11bp,39.981bp);
  \draw (186.5bp,50.5bp) node {$(0,3](3,6]$};
  \draw [->,very thick] (35.954bp,19.594bp) .. controls (50.163bp,21.027bp) and (70.465bp,23.456bp)  .. (88.0bp,27.0bp) .. controls (90.861bp,27.578bp) and (93.829bp,28.262bp)  .. (106.54bp,31.61bp);
  \draw (71.0bp,34.5bp) node {$(0,2]$};
  \draw [->,very thick] (237.09bp,23.41bp) .. controls (230.66bp,17.123bp) and (222.18bp,10.324bp)  .. (213.0bp,7.0bp) .. controls (190.85bp,-1.0189bp) and (182.15bp,-1.0189bp)  .. (160.0bp,7.0bp) .. controls (154.12bp,9.1291bp) and (148.52bp,12.685bp)  .. (135.91bp,23.41bp);
  \draw (186.5bp,22.5bp) node {$(0,4]$};
\begin{scope}
  \definecolor{strokecol}{rgb}{0.0,0.0,0.0};
  \pgfsetstrokecolor{strokecol}
  \draw [very thick] (18.0bp,72.0bp) ellipse (18.0bp and 18.0bp);
  \draw (18.0bp,72.0bp) node {$A$};
   \draw (-26bp,79.0bp) node {$\lambda_{A,c_l(h,t)} $};
\end{scope}
\begin{scope}
  \definecolor{strokecol}{rgb}{0.0,0.0,0.0};
  \pgfsetstrokecolor{strokecol}
  \draw [very thick] (124.0bp,37.0bp) ellipse (18.0bp and 18.0bp);
  \draw (124.0bp,37.0bp) node {$C$};
  \draw (124.0bp,65.3bp) node {$\lambda_{C,c_l(h,t)} $};
\end{scope}
\begin{scope}
  \definecolor{strokecol}{rgb}{0.0,0.0,0.0};
  \pgfsetstrokecolor{strokecol}
  \draw [very thick] (249.0bp,37.0bp) ellipse (18.0bp and 18.0bp);
  \draw (249.0bp,37.0bp) node {$D$};
    \draw (249.0bp,65.3bp) node {$\lambda_{D,c_l(h,t)} $};
\end{scope}
\begin{scope}
  \definecolor{strokecol}{rgb}{0.0,0.0,0.0};
  \pgfsetstrokecolor{strokecol}
  \draw [very thick] (18.0bp,18.0bp) ellipse (18.0bp and 18.0bp);
  \draw (18.0bp,18.0bp) node {$B$};
    \draw (-10.0bp,25.3bp) node {$ \lambda_B $};
\end{scope}
\end{tikzpicture}

%% file: figures/time_line.tex
\begin{tikzpicture}[x=1pt,y=1pt, scale = 0.9]
\definecolor{fillColor}{RGB}{255,255,255}
\path[use as bounding box,fill=fillColor,fill opacity=0.00] (0,0) rectangle (469.75, 86.72);
\begin{scope}
\path[clip] (  0.00,  0.00) rectangle (469.75, 86.72);
\definecolor{drawColor}{RGB}{255,255,255}
\definecolor{fillColor}{RGB}{255,255,255}

\path[draw=drawColor,line width= 0.6pt,line join=round,line cap=round,fill=fillColor] (  0.00,  0.00) rectangle (469.75, 86.72);
\end{scope}
\begin{scope}
\path[clip] (  8.25,  8.25) rectangle (409.97, 81.22);
\definecolor{fillColor}{RGB}{255,255,255}

\path[fill=fillColor] (  8.25,  8.25) rectangle (409.97, 81.22);
\definecolor{drawColor}{RGB}{27,158,119}

\path[draw=drawColor,line width= 0.2pt,line join=round] ( 90.96, 55.79) -- (119.60, 55.79);

\path[draw=drawColor,line width= 0.2pt,line join=round] (114.67, 52.95) --
	(119.60, 55.79) --
	(114.67, 58.64);
\definecolor{drawColor}{RGB}{217,95,2}

\path[draw=drawColor,line width= 0.2pt,line join=round] ( 69.47, 66.85) -- ( 98.12, 66.85);

\path[draw=drawColor,line width= 0.2pt,line join=round] ( 93.19, 64.01) --
	( 98.12, 66.85) --
	( 93.19, 69.70);
\definecolor{drawColor}{RGB}{27,158,119}

\path[draw=drawColor,line width= 0.2pt,line join=round] (148.24, 55.79) -- (176.88, 55.79);

\path[draw=drawColor,line width= 0.2pt,line join=round] (171.96, 52.95) --
	(176.88, 55.79) --
	(171.96, 58.64);
\definecolor{drawColor}{RGB}{117,112,179}

\path[draw=drawColor,line width= 0.2pt,line join=round] (162.56, 77.91) -- (191.21, 77.91);

\path[draw=drawColor,line width= 0.2pt,line join=round] (186.28, 75.06) --
	(191.21, 77.91) --
	(186.28, 80.75);
\definecolor{drawColor}{RGB}{217,95,2}

\path[draw=drawColor,line width= 0.2pt,line join=round] (212.69, 66.85) -- (241.33, 66.85);

\path[draw=drawColor,line width= 0.2pt,line join=round] (236.40, 64.01) --
	(241.33, 66.85) --
	(236.40, 69.70);
\definecolor{drawColor}{RGB}{27,158,119}

\path[draw=drawColor,line width= 0.2pt,line join=round] (284.30, 55.79) -- (312.94, 55.79);

\path[draw=drawColor,line width= 0.2pt,line join=round] (308.01, 52.95) --
	(312.94, 55.79) --
	(308.01, 58.64);
\definecolor{drawColor}{RGB}{117,112,179}

\path[draw=drawColor,line width= 0.2pt,line join=round] (305.78, 77.91) -- (334.42, 77.91);

\path[draw=drawColor,line width= 0.2pt,line join=round] (329.49, 75.06) --
	(334.42, 77.91) --
	(329.49, 80.75);
\definecolor{drawColor}{RGB}{217,95,2}

\path[draw=drawColor,line width= 0.2pt,line join=round] (341.58, 66.85) -- (370.22, 66.85);

\path[draw=drawColor,line width= 0.2pt,line join=round] (365.30, 64.01) --
	(370.22, 66.85) --
	(365.30, 69.70);
\definecolor{drawColor}{RGB}{27,158,119}

\path[draw=drawColor,line width= 0.2pt,line join=round] (119.60, 55.79) -- ( 90.96, 55.79);

\path[draw=drawColor,line width= 0.2pt,line join=round] ( 95.88, 58.64) --
	( 90.96, 55.79) --
	( 95.88, 52.95);
\definecolor{drawColor}{RGB}{217,95,2}

\path[draw=drawColor,line width= 0.2pt,line join=round] ( 98.12, 66.85) -- ( 69.47, 66.85);

\path[draw=drawColor,line width= 0.2pt,line join=round] ( 74.40, 69.70) --
	( 69.47, 66.85) --
	( 74.40, 64.01);
\definecolor{drawColor}{RGB}{27,158,119}

\path[draw=drawColor,line width= 0.2pt,line join=round] (176.88, 55.79) -- (148.24, 55.79);

\path[draw=drawColor,line width= 0.2pt,line join=round] (153.17, 58.64) --
	(148.24, 55.79) --
	(153.17, 52.95);
\definecolor{drawColor}{RGB}{117,112,179}

\path[draw=drawColor,line width= 0.2pt,line join=round] (191.21, 77.91) -- (162.56, 77.91);

\path[draw=drawColor,line width= 0.2pt,line join=round] (167.49, 80.75) --
	(162.56, 77.91) --
	(167.49, 75.06);
\definecolor{drawColor}{RGB}{217,95,2}

\path[draw=drawColor,line width= 0.2pt,line join=round] (241.33, 66.85) -- (212.69, 66.85);

\path[draw=drawColor,line width= 0.2pt,line join=round] (217.62, 69.70) --
	(212.69, 66.85) --
	(217.62, 64.01);
\definecolor{drawColor}{RGB}{27,158,119}

\path[draw=drawColor,line width= 0.2pt,line join=round] (312.94, 55.79) -- (284.30, 55.79);

\path[draw=drawColor,line width= 0.2pt,line join=round] (289.22, 58.64) --
	(284.30, 55.79) --
	(289.22, 52.95);
\definecolor{drawColor}{RGB}{117,112,179}

\path[draw=drawColor,line width= 0.2pt,line join=round] (334.42, 77.91) -- (305.78, 77.91);

\path[draw=drawColor,line width= 0.2pt,line join=round] (310.71, 80.75) --
	(305.78, 77.91) --
	(310.71, 75.06);
\definecolor{drawColor}{RGB}{217,95,2}

\path[draw=drawColor,line width= 0.2pt,line join=round] (370.22, 66.85) -- (341.58, 66.85);

\path[draw=drawColor,line width= 0.2pt,line join=round] (346.51, 69.70) --
	(341.58, 66.85) --
	(346.51, 64.01);
\definecolor{drawColor}{RGB}{27,158,119}

\path[draw=drawColor,draw opacity=0.20,line width= 0.2pt,dash pattern=on 4pt off 4pt ,line join=round] ( 33.67, 55.79) -- (391.71, 55.79);
\definecolor{drawColor}{RGB}{217,95,2}

\path[draw=drawColor,draw opacity=0.20,line width= 0.2pt,dash pattern=on 4pt off 4pt ,line join=round] ( 33.67, 66.85) -- (391.71, 66.85);
\definecolor{drawColor}{RGB}{27,158,119}

\path[draw=drawColor,draw opacity=0.20,line width= 0.2pt,dash pattern=on 4pt off 4pt ,line join=round] ( 33.67, 55.79) -- (391.71, 55.79);
\definecolor{drawColor}{RGB}{117,112,179}

\path[draw=drawColor,draw opacity=0.20,line width= 0.2pt,dash pattern=on 4pt off 4pt ,line join=round] ( 33.67, 77.91) -- (391.71, 77.91);
\definecolor{drawColor}{RGB}{217,95,2}

\path[draw=drawColor,draw opacity=0.20,line width= 0.2pt,dash pattern=on 4pt off 4pt ,line join=round] ( 33.67, 66.85) -- (391.71, 66.85);
\definecolor{drawColor}{RGB}{27,158,119}

\path[draw=drawColor,draw opacity=0.20,line width= 0.2pt,dash pattern=on 4pt off 4pt ,line join=round] ( 33.67, 55.79) -- (391.71, 55.79);
\definecolor{drawColor}{RGB}{117,112,179}

\path[draw=drawColor,draw opacity=0.20,line width= 0.2pt,dash pattern=on 4pt off 4pt ,line join=round] ( 33.67, 77.91) -- (391.71, 77.91);
\definecolor{drawColor}{RGB}{217,95,2}

\path[draw=drawColor,draw opacity=0.20,line width= 0.2pt,dash pattern=on 4pt off 4pt ,line join=round] ( 33.67, 66.85) -- (391.71, 66.85);
\definecolor{drawColor}{RGB}{0,0,0}

\path[draw=drawColor,line width= 0.6pt,line join=round] ( 33.67, 33.68) -- (391.71, 33.68);

\path[draw=drawColor,line width= 0.6pt,line join=round] ( 33.67, 33.68) -- (391.71, 33.68);

\path[draw=drawColor,line width= 0.6pt,line join=round] ( 33.67, 33.68) -- (391.71, 33.68);

\path[draw=drawColor,line width= 0.6pt,line join=round] ( 33.67, 33.68) -- (391.71, 33.68);

\path[draw=drawColor,line width= 0.6pt,line join=round] ( 33.67, 33.68) -- (391.71, 33.68);

\path[draw=drawColor,line width= 0.6pt,line join=round] ( 33.67, 33.68) -- (391.71, 33.68);

\path[draw=drawColor,line width= 0.6pt,line join=round] ( 33.67, 33.68) -- (391.71, 33.68);

\path[draw=drawColor,line width= 0.6pt,line join=round] ( 33.67, 33.68) -- (391.71, 33.68);
\definecolor{fillColor}{RGB}{27,158,119}

\path[fill=fillColor] ( 90.96, 33.68) circle (  4.10);
\definecolor{fillColor}{RGB}{217,95,2}

\path[fill=fillColor] ( 69.47, 40.06) --
	( 75.00, 30.49) --
	( 63.95, 30.49) --
	cycle;
\definecolor{fillColor}{RGB}{27,158,119}

\path[fill=fillColor] (148.24, 33.68) circle (  4.10);
\definecolor{fillColor}{RGB}{117,112,179}

\path[fill=fillColor] (158.46, 29.58) --
	(166.67, 29.58) --
	(166.67, 37.78) --
	(158.46, 37.78) --
	cycle;
\definecolor{fillColor}{RGB}{217,95,2}

\path[fill=fillColor] (212.69, 40.06) --
	(218.22, 30.49) --
	(207.16, 30.49) --
	cycle;
\definecolor{fillColor}{RGB}{27,158,119}

\path[fill=fillColor] (284.30, 33.68) circle (  4.10);
\definecolor{fillColor}{RGB}{117,112,179}

\path[fill=fillColor] (301.67, 29.58) --
	(309.88, 29.58) --
	(309.88, 37.78) --
	(301.67, 37.78) --
	cycle;
\definecolor{fillColor}{RGB}{217,95,2}

\path[fill=fillColor] (341.58, 40.06) --
	(347.11, 30.49) --
	(336.05, 30.49) --
	cycle;

\node[text=drawColor,anchor=base,inner sep=0pt, outer sep=0pt, scale=  1.00] at ( 90.96,  8.14) {4};

\node[text=drawColor,anchor=base,inner sep=0pt, outer sep=0pt, scale=  1.00] at ( 69.47,  8.14) {2.5};

\node[text=drawColor,anchor=base,inner sep=0pt, outer sep=0pt, scale=  1.00] at (148.24,  8.14) {8};

\node[text=drawColor,anchor=base,inner sep=0pt, outer sep=0pt, scale=  1.00] at (162.56,  8.14) {9};

\node[text=drawColor,anchor=base,inner sep=0pt, outer sep=0pt, scale=  1.00] at (212.69,  8.14) {12.5};

\node[text=drawColor,anchor=base,inner sep=0pt, outer sep=0pt, scale=  1.00] at (284.30,  8.14) {17.5};

\node[text=drawColor,anchor=base,inner sep=0pt, outer sep=0pt, scale=  1.00] at (305.78,  8.14) {19};

\node[text=drawColor,anchor=base,inner sep=0pt, outer sep=0pt, scale=  1.00] at (341.58,  8.14) {21.5};

\node[text=drawColor,anchor=base,inner sep=0pt, outer sep=0pt, scale=  1.00] at ( 33.67,  8.14) {0};

\node[text=drawColor,anchor=base,inner sep=0pt, outer sep=0pt, scale=  1.00] at ( 33.67,  8.14) {0};

\node[text=drawColor,anchor=base,inner sep=0pt, outer sep=0pt, scale=  1.00] at ( 33.67,  8.14) {0};

\node[text=drawColor,anchor=base,inner sep=0pt, outer sep=0pt, scale=  1.00] at ( 33.67,  8.14) {0};

\node[text=drawColor,anchor=base,inner sep=0pt, outer sep=0pt, scale=  1.00] at ( 33.67,  8.14) {0};

\node[text=drawColor,anchor=base,inner sep=0pt, outer sep=0pt, scale=  1.00] at ( 33.67,  8.14) {0};

\node[text=drawColor,anchor=base,inner sep=0pt, outer sep=0pt, scale=  1.00] at ( 33.67,  8.14) {0};

\node[text=drawColor,anchor=base,inner sep=0pt, outer sep=0pt, scale=  1.00] at ( 33.67,  8.14) {0};

\node[text=drawColor,anchor=base,inner sep=0pt, outer sep=0pt, scale=  1.00] at (391.71,  8.14) {25};

\node[text=drawColor,anchor=base,inner sep=0pt, outer sep=0pt, scale=  1.00] at (391.71,  8.14) {25};

\node[text=drawColor,anchor=base,inner sep=0pt, outer sep=0pt, scale=  1.00] at (391.71,  8.14) {25};

\node[text=drawColor,anchor=base,inner sep=0pt, outer sep=0pt, scale=  1.00] at (391.71,  8.14) {25};

\node[text=drawColor,anchor=base,inner sep=0pt, outer sep=0pt, scale=  1.00] at (391.71,  8.14) {25};

\node[text=drawColor,anchor=base,inner sep=0pt, outer sep=0pt, scale=  1.00] at (391.71,  8.14) {25};

\node[text=drawColor,anchor=base,inner sep=0pt, outer sep=0pt, scale=  1.00] at (391.71,  8.14) {25};

\node[text=drawColor,anchor=base,inner sep=0pt, outer sep=0pt, scale=  1.00] at (391.71,  8.14) {25};
\definecolor{drawColor}{RGB}{255,255,255}

\path[draw=drawColor,line width= 0.3pt,line join=round] ( 26.51, 33.68) -- ( 26.51, 77.91);

\path[draw=drawColor,line width= 0.3pt,line join=round] ( 26.51, 33.68) -- ( 26.51, 77.91);

\path[draw=drawColor,line width= 0.3pt,line join=round] ( 26.51, 33.68) -- ( 26.51, 77.91);

\path[draw=drawColor,line width= 0.3pt,line join=round] ( 26.51, 33.68) -- ( 26.51, 77.91);

\path[draw=drawColor,line width= 0.3pt,line join=round] ( 26.51, 33.68) -- ( 26.51, 77.91);

\path[draw=drawColor,line width= 0.3pt,line join=round] ( 26.51, 33.68) -- ( 26.51, 77.91);

\path[draw=drawColor,line width= 0.3pt,line join=round] ( 26.51, 33.68) -- ( 26.51, 77.91);

\path[draw=drawColor,line width= 0.3pt,line join=round] ( 26.51, 33.68) -- ( 26.51, 77.91);
\definecolor{drawColor}{RGB}{0,0,0}

\path[draw=drawColor,line width= 0.3pt,line join=round] ( 33.67, 29.26) -- ( 33.67, 38.10);

\path[draw=drawColor,line width= 0.3pt,line join=round] ( 33.67, 29.26) -- ( 33.67, 38.10);

\path[draw=drawColor,line width= 0.3pt,line join=round] ( 33.67, 29.26) -- ( 33.67, 38.10);

\path[draw=drawColor,line width= 0.3pt,line join=round] ( 33.67, 29.26) -- ( 33.67, 38.10);

\path[draw=drawColor,line width= 0.3pt,line join=round] ( 33.67, 29.26) -- ( 33.67, 38.10);

\path[draw=drawColor,line width= 0.3pt,line join=round] ( 33.67, 29.26) -- ( 33.67, 38.10);

\path[draw=drawColor,line width= 0.3pt,line join=round] ( 33.67, 29.26) -- ( 33.67, 38.10);

\path[draw=drawColor,line width= 0.3pt,line join=round] ( 33.67, 29.26) -- ( 33.67, 38.10);

\path[draw=drawColor,line width= 0.3pt,line join=round] (391.71, 29.26) -- (391.71, 38.10);

\path[draw=drawColor,line width= 0.3pt,line join=round] (391.71, 29.26) -- (391.71, 38.10);

\path[draw=drawColor,line width= 0.3pt,line join=round] (391.71, 29.26) -- (391.71, 38.10);

\path[draw=drawColor,line width= 0.3pt,line join=round] (391.71, 29.26) -- (391.71, 38.10);

\path[draw=drawColor,line width= 0.3pt,line join=round] (391.71, 29.26) -- (391.71, 38.10);

\path[draw=drawColor,line width= 0.3pt,line join=round] (391.71, 29.26) -- (391.71, 38.10);

\path[draw=drawColor,line width= 0.3pt,line join=round] (391.71, 29.26) -- (391.71, 38.10);

\path[draw=drawColor,line width= 0.3pt,line join=round] (391.71, 29.26) -- (391.71, 38.10);
\end{scope}
\begin{scope}
\path[clip] (  0.00,  0.00) rectangle (469.75, 86.72);
\definecolor{fillColor}{RGB}{255,255,255}

\path[fill=fillColor] (420.97,  9.95) rectangle (464.25, 79.53);
\end{scope}
\begin{scope}
\path[clip] (  0.00,  0.00) rectangle (469.75, 86.72);
\definecolor{drawColor}{RGB}{0,0,0}

\node[text=drawColor,anchor=base west,inner sep=0pt, outer sep=0pt, scale=  1.10] at (426.47, 65.38) {Events};
\end{scope}
\begin{scope}
\path[clip] (  0.00,  0.00) rectangle (469.75, 86.72);
\definecolor{drawColor}{RGB}{27,158,119}

\path[draw=drawColor,line width= 0.2pt,line join=round] (427.91, 51.58) -- (439.47, 51.58);

\path[draw=drawColor,line width= 0.2pt,line join=round] (434.55, 48.74) --
	(439.47, 51.58) --
	(434.55, 54.43);
\end{scope}
\begin{scope}
\path[clip] (  0.00,  0.00) rectangle (469.75, 86.72);
\definecolor{drawColor}{RGB}{27,158,119}

\path[draw=drawColor,line width= 0.2pt,line join=round] (427.91, 51.58) -- (439.47, 51.58);

\path[draw=drawColor,line width= 0.2pt,line join=round] (434.55, 48.74) --
	(439.47, 51.58) --
	(434.55, 54.43);
\end{scope}
\begin{scope}
\path[clip] (  0.00,  0.00) rectangle (469.75, 86.72);
\definecolor{drawColor}{RGB}{27,158,119}

\path[draw=drawColor,draw opacity=0.20,line width= 0.2pt,dash pattern=on 4pt off 4pt ,line join=round] (427.91, 51.58) -- (439.47, 51.58);
\end{scope}
\begin{scope}
\path[clip] (  0.00,  0.00) rectangle (469.75, 86.72);
\definecolor{fillColor}{RGB}{27,158,119}

\path[fill=fillColor] (433.69, 51.58) circle (  4.10);
\end{scope}
\begin{scope}
\path[clip] (  0.00,  0.00) rectangle (469.75, 86.72);
\definecolor{drawColor}{RGB}{217,95,2}

\path[draw=drawColor,line width= 0.2pt,line join=round] (427.91, 37.13) -- (439.47, 37.13);

\path[draw=drawColor,line width= 0.2pt,line join=round] (434.55, 34.28) --
	(439.47, 37.13) --
	(434.55, 39.98);
\end{scope}
\begin{scope}
\path[clip] (  0.00,  0.00) rectangle (469.75, 86.72);
\definecolor{drawColor}{RGB}{217,95,2}

\path[draw=drawColor,line width= 0.2pt,line join=round] (427.91, 37.13) -- (439.47, 37.13);

\path[draw=drawColor,line width= 0.2pt,line join=round] (434.55, 34.28) --
	(439.47, 37.13) --
	(434.55, 39.98);
\end{scope}
\begin{scope}
\path[clip] (  0.00,  0.00) rectangle (469.75, 86.72);
\definecolor{drawColor}{RGB}{217,95,2}

\path[draw=drawColor,draw opacity=0.20,line width= 0.2pt,dash pattern=on 4pt off 4pt ,line join=round] (427.91, 37.13) -- (439.47, 37.13);
\end{scope}
\begin{scope}
\path[clip] (  0.00,  0.00) rectangle (469.75, 86.72);
\definecolor{fillColor}{RGB}{217,95,2}

\path[fill=fillColor] (433.69, 43.51) --
	(439.22, 33.94) --
	(428.17, 33.94) --
	cycle;
\end{scope}
\begin{scope}
\path[clip] (  0.00,  0.00) rectangle (469.75, 86.72);
\definecolor{drawColor}{RGB}{117,112,179}

\path[draw=drawColor,line width= 0.2pt,line join=round] (427.91, 22.68) -- (439.47, 22.68);

\path[draw=drawColor,line width= 0.2pt,line join=round] (434.55, 19.83) --
	(439.47, 22.68) --
	(434.55, 25.52);
\end{scope}
\begin{scope}
\path[clip] (  0.00,  0.00) rectangle (469.75, 86.72);
\definecolor{drawColor}{RGB}{117,112,179}

\path[draw=drawColor,line width= 0.2pt,line join=round] (427.91, 22.68) -- (439.47, 22.68);

\path[draw=drawColor,line width= 0.2pt,line join=round] (434.55, 19.83) --
	(439.47, 22.68) --
	(434.55, 25.52);
\end{scope}
\begin{scope}
\path[clip] (  0.00,  0.00) rectangle (469.75, 86.72);
\definecolor{drawColor}{RGB}{117,112,179}

\path[draw=drawColor,draw opacity=0.20,line width= 0.2pt,dash pattern=on 4pt off 4pt ,line join=round] (427.91, 22.68) -- (439.47, 22.68);
\end{scope}
\begin{scope}
\path[clip] (  0.00,  0.00) rectangle (469.75, 86.72);
\definecolor{fillColor}{RGB}{117,112,179}

\path[fill=fillColor] (429.59, 18.57) --
	(437.80, 18.57) --
	(437.80, 26.78) --
	(429.59, 26.78) --
	cycle;
\end{scope}
\begin{scope}
\path[clip] (  0.00,  0.00) rectangle (469.75, 86.72);
\definecolor{drawColor}{RGB}{0,0,0}

\node[text=drawColor,anchor=base west,inner sep=0pt, outer sep=0pt, scale=  0.88] at (446.42, 48.55) {A};
\end{scope}
\begin{scope}
\path[clip] (  0.00,  0.00) rectangle (469.75, 86.72);
\definecolor{drawColor}{RGB}{0,0,0}

\node[text=drawColor,anchor=base west,inner sep=0pt, outer sep=0pt, scale=  0.88] at (446.42, 34.10) {B};
\end{scope}
\begin{scope}
\path[clip] (  0.00,  0.00) rectangle (469.75, 86.72);
\definecolor{drawColor}{RGB}{0,0,0}

\node[text=drawColor,anchor=base west,inner sep=0pt, outer sep=0pt, scale=  0.88] at (446.42, 19.65) {C};
\end{scope}
\end{tikzpicture}

%% file: chapters/03_methods.tex
\section{Experiments}

To the best of my knowledge, TGEMs are so far only theoretically covered in literature \citep{Gunawardana2016, Antakly19, Monvoisin2019} and neither synthetic nor real-world data have been modeled yet with TGEMs. Moreover, a relevant question - the choice of the default horizon - has not received any dedication. This report aims to fill this gap. Hence, I  conduct a comprehensive benchmark on synthetic data. This will allow to evaluate performance of the different heuristics that I proposed in section \ref{Horizon Choice}.

To test the capability of learning algorithms in the area of graphical models, conducting  benchmarks on synthetic data sets is a very common approach \citep{pnetwork2005,Tsamardinos2006,Weiss2013, Bhattacharjya2018}. However, unlike for Bayesian Networks\footnote{The bn-learn package of \citet{Scutarie} and its repository \url{https://www.bnlearn.com/bnrepository/} provides several gold standard models.}, there exist no such pre-defined models for TGEMs.  Thus, I will create random TGEMs  according to the Erdős–Rényi model for random graph generation \citep{Erdos:1960}. From each of these TGEMs, I will generate synthetic data sets as described in section \ref{Sampling}, re-learn TGEMs from these data sets and finally measure its distance to the data-generating TGEM. As discussed in section \ref{Distance Measure}, I will apply the refined definition of the \textit{elementary distance}. Additionally, the $F1$-score will be reported considering whether a true dependency is learned or not. Third, the capability of learning edges with different temporal ranges will be examined by reporting the distance per horizon aggregated over all graphs. This procedure will allow to draw conclusions about (1) the different heuristics to determine the default horizon, (2) the ability to learn temporal dependencies of different ranges.  

To determine the data-generating TGEMs, various parameters need to be set: the number of nodes $|\mathcal{L}| $, the set of edges $E$, its timescales $\mathcal{T}$ incorporating the range of the temporal dependencies, as well as the (conditional) intensity functions  
$\lambda_{l,c_l(h,t)}$. Moreover, the sampled time units (i.e., the length of the data set) need to be defined.

\input{figures/benchmark_params.tex}

The relevant parameters for the benchmark are shown in Table \ref{tab:benchmark}. 
To allow general conclusion, I will consider TGEMs with different properties. Foremost, this is the size of the graph. During the benchmark, I will consider TGEMs with small and moderate size:  $|\mathcal{L}| \in M $. Edges are randomly drawn with a constant probability $d \in D = \{0.1, 0.2 \}$. This allows to vary the complexity. Parameter $d$ can be understood as \textit{density} of the graph indicating how many edges with respect to the total number of possible edges are expected. For each existing edge an initial timescale with one interval is set. Its horizon  is randomly chosen from $H$ allowing to model dependencies of various duration. One might understand a time unit as hour, thus, the horizons can represent a dependency between 30 minutes and one day. 
 
Further, splits or extends might be applied. Therefore, I draw from a geometric distribution $P(i|p) = p  (1 - p)^i$ with $p = 0.85$ the number of additional modifications on the corresponding timescale. The number of additional modifications is randomly assigned to splits or extends which are consecutively executed\footnote{Extends can only be applied to the last  interval of the timescale, the split however, will be again randomly assigned to one of the intervals of the corresponding timescale}. According to the behaviour of the geometric distribution, this parametrization will yield many timescales containing a single interval and only few timescales with multiple intervals. Moreover, each node is restricted to an in-degree of two and moreover  maximal four intervals on all incoming edges together\footnote{Number of parameters grow exponentially to the power of $2$. Accordingly, allowing more than 4 intervals requires to provide at least 32 parameters per node}. The rates of the (conditional) intensity functions are randomly picked from $\Lambda$ allowing to model various patterns ( expected event occurrence between approximately every $1.5$th  and every $100$th time unit).

For the benchmark, I span a grid containing all possible combinations of number of nodes and densities ($M \times D$). For each cell in this grid, I create $100$ random TGEMs following the afore-mentioned procedure. One example graph for the setting $|\mathcal{L}| = 5$, $d = 0.2 $  is depicted in Figure \ref{fig:random_graph} in \nameref{ch:ApppendigxA}. For each TGEM, I generate data sets of different lengths $t \in T$. This yields $ 3 \times 2 \times 100 \times 5 = 3,000 $ different data sets. 

To test the choice of the default horizon, I learn each data set with the \textit{proximal heuristic} and with the naive \textit{quantile heuristic} for various quantiles $q \in \{ 0.05, 0.25, 0.5, 0.75, 0.95 \}$. Overall, $18,000$ models are calculated and compared.

%% file: figures/benchmark_params.tex
\begin{table}[htb]
\centering
\fontsize{10}{12}\selectfont
\caption[Benchmark parametrization]{Parametrization of data generating TGEMs in the benchmark}
\label{tab:benchmark}
\begin{tabular}{l|ccc}
\toprule
\textbf{Parameter} & \textbf{Symbol} & \textbf{Values}              \\ \midrule
Number of nodes    & $M$                                 & $\{5, 10, 15 \}$                                 \\
Density of graph   & $D$                                 & $\{0.1, 0.2\}$                                   \\
Sampled time units & $T$                                 & $\{500, 1000, 2000, 4000, 8000 \} $              \\
Initial Horizons   & $H$                                 & $\{1, 2,4, 8, 16, 24  \}$                        \\
Intensity rates    & $\Lambda$                           & $\{ 0.01, 0.02, 0.04, 0.08, 0.16, 0.32, 0.64 \}$ \\ \bottomrule
\end{tabular}
\end{table}

%% file: chapters/04_analysis.tex
\section{Results}

\begin{figure}[htb]
\centering
\input{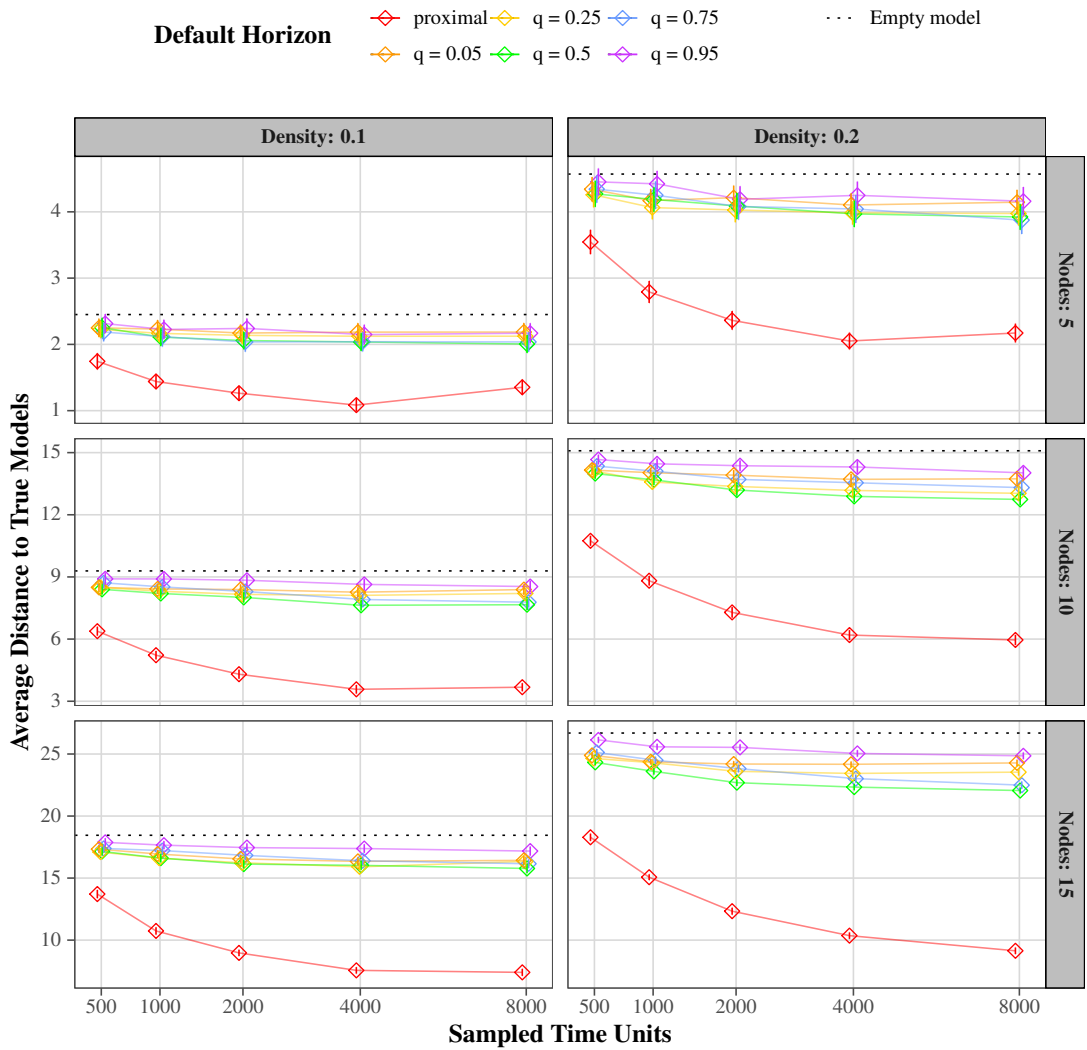}
\caption[Benchmark results: average distance per heuristic]{Benchmark results for different default horizon heuristics: average distance between the learned model and $100$  data generating TGEMs for various properties (Nodes, Density), and data sets of different sizes}
		\label{plot:results_benchmark}
\end{figure}

Figure \ref{plot:results_benchmark} provides a comprehensive overview for the results of the benchmark. It depicts the distance between the data generating models and the learned models for the six employed heuristics with respect to the size of the data sets. Further, a hypothetical distance for a weak baseline model without any edges is mapped. The reported distances are averages over $100$ TGEMs and error bars indicate the standard error. Each subplot represents a different \textit{setting} in which the data generating models vary in their number of nodes (rows) and their density (columns) as described in section  \ref{ch:synthetic}. 

First of all, the proposed algorithm of \citet{Gunawardana2016} is able to learn the interdependencies from event streams and the quality generally increases with larger data sets.  On average, it performs better than an weak baseline model treating each event independently (i.e., a TGEM without edges). As the data generating graphs are expected to have  $|E| = d * |\mathcal{M}|^2$ edges, the distance of an empty baseline model would coincide with $|E|$. However, due to the thresholds set during the random TGEM generation, the empirical number of edges is lower. For instance, one might expect $20$ edges for a graph of $10$ nodes and a density of $0.2$ but on average only $15$ are generated. Consequently, the distance of the empty model to the data generating graph is set to $15$.

With respect to the choice of the default horizon, there is no doubt about the superiority of the \textit{proximal heuristic}. It outperforms the naive quantile approach regardless of the choice of $q$. In each constellation, learning with the  \textit{proximal heuristic} yields models that are considerably  closer to the true models than the other options. Secondly, it benefits stronger from increasing data set sizes. Whereas the performance of the different \textit{quantile heuristics} decreases only slightly with more data, the proximal heuristic has a steeper learning curve. For instance, the average distance between data generating models with $15$ nodes and a density of $0.1$ to the  models learned by the \textit{proximal heuristic}  equals $13.71$ for a data set containing events for $500$ time units. For $ 2,000$ time units, however, the average distance is approximately $9$. On the contrary, for the different \textit{quantile heuristics} the improvement is  rather little from on average $17$ to $16$. The detailed numbers can be found in Table \ref{tab:benchmark_dist} in \nameref{ch:ApppendigxA}. 

Among the \textit{quantile heuristics}, the choice of the median ($q = 0.5$) followed by first quartile ($q = 0.25$) tend to learn models closest to the true ones. However, the differences to other choices of $q$ are diminutive compared to their differences to the performance of the \textit{proximal heuristic}. Globally, the gap between the various \textit{quantile heuristics} and the empty model is rather low indicating a weak capability of inferring the right temporal range and consecutively dependencies. 

The applied distance measure considered not only the existence of an edge but also the differences between respective timescales. From a pure qualitative perspective one might ask whether a true dependency between two nodes in the data generating model is found, hence perceiving it as a binary classification. This allows to investigate the two different kind of errors (\textit{false positives} and \textit{false negatives}) that can be made during the model estimation between which the distance measure did not distinguish. The $F1$-score as harmonic mean between precision and recall penalizes classifiers that tend to favor one of the errors. For instance, the baseline model without edges could never exhibit false positives (as it never assumes any dependency) but only false negatives. Thus, the $F1$-score would equal $0$. 

Overall, the results for the $F1$-score correspond to those for the distance measure. The \textit{proximal heuristic}
yields by far the highest $F1$-score for each constellation, regularly exceeding $0.7$. However, it allows a clearer distinction between the different operationalizations of the \textit{quantile heuristics}. With large data sets ($8,000$ time units), the median heuristic exceeds a $F1$-score of $0.6$. The same holds for $q = 0.25$. On the contrary, extreme values for $q$ ($0.95, 0.05$) yield relatively low values with approximately $0.4 - 0.5$. The respective Figure \ref{plot:results_benchmark_f1} and Table \ref{tab:F1_score} with the exact results can be found in \nameref{ch:ApppendigxA}. 

Table \ref{tab:avg_events} provides summary statistics of the event occurrences for the different data sets. On average, the nodes with the fewest occurrences were found $10$ times within $500$ sampled time units whereas the median occurrences equaled $69$ and the maximum $265$. Conversely, in the largest data sets, the node with the fewest occurrences is found $164$ times on average. The node with the most occurrences is found  $4,209$ times on average.

\input{figures/average_events.tex}

\begin{figure}[htb]
\centering
\input{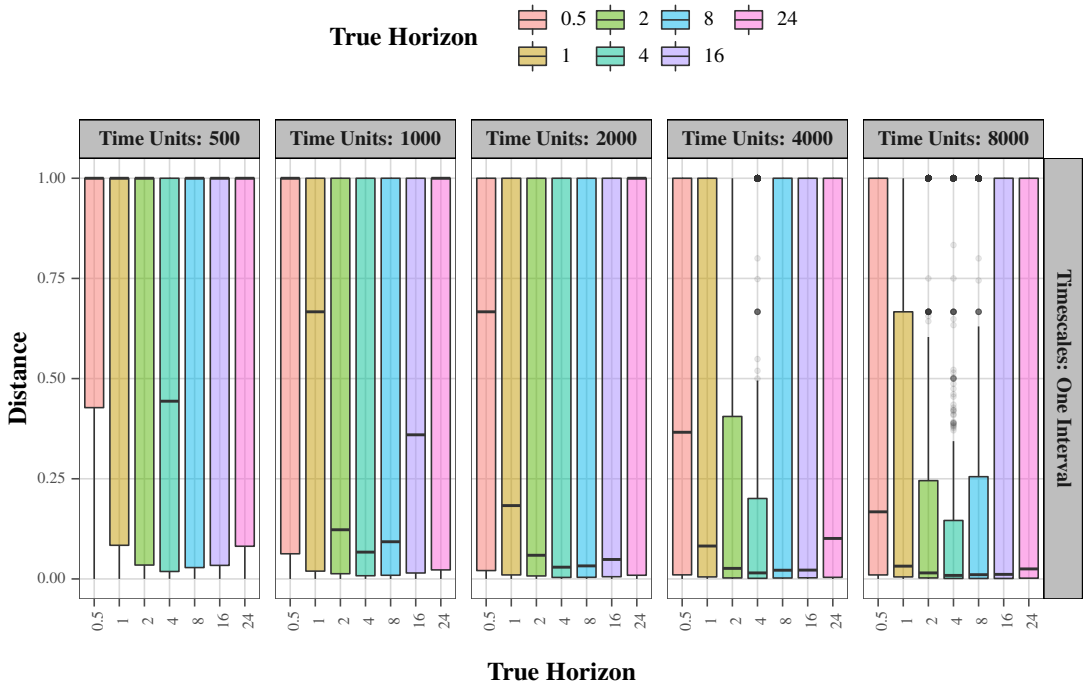}
\caption[Distance per horizon for singular timescales]{Box plots for the distance between edges with a single interval on their timescales in the data generating TGEMs to the respective edges in the learned TGEMs by the proximal heuristic with respect to the true horizon}
		\label{plot:results_single}
\end{figure}

To test the robustness of TGEMs with respect to their ability to learn temporal dependencies of different lengths, I analyzed the distances per  horizon of the edge in the true model. As the \textit{proximal heuristic} clearly outperformed the other heuristics, I only consider  models that have been learned with this approach in my analysis\footnote{This applies to all subsequent approaches with TGEMs.}. Figure \ref{plot:results_single} displays the distributions of the distances between existing edges in the data generating model and their possibly learned equivalents with respect to the true duration and restricted to edges with a single interval on their timescale. Each subplot accounts for a different size of the data sets. Globally, TGEMs are able to detect temporal dependencies of different length. However, it requires a given certain of data. For instance, from data sets with  $500$ sampled time units the median distance is $1$ for edges with each horizon (except $4$). Thus, in $50 \%$ of the cases an edge is not even found. For data sets containing $2,000$ sampled time units however, dependencies of medium temporal ranges are learned quite reliably. The median distance for edges with horizons between $1$ time unit and $16$ time units is below $0.2$. For the two extreme choices for the horizon ($0.5, 24$) the median distance is notably higher with $0.65$ and $1$ respectively. For the largest data sets in the benchmark, the median distances for edges with each horizon are  below $0.2$. Nonetheless, medium ranged horizon exhibit a lower variation and converge faster.

\begin{figure}[htb]
\centering
\input{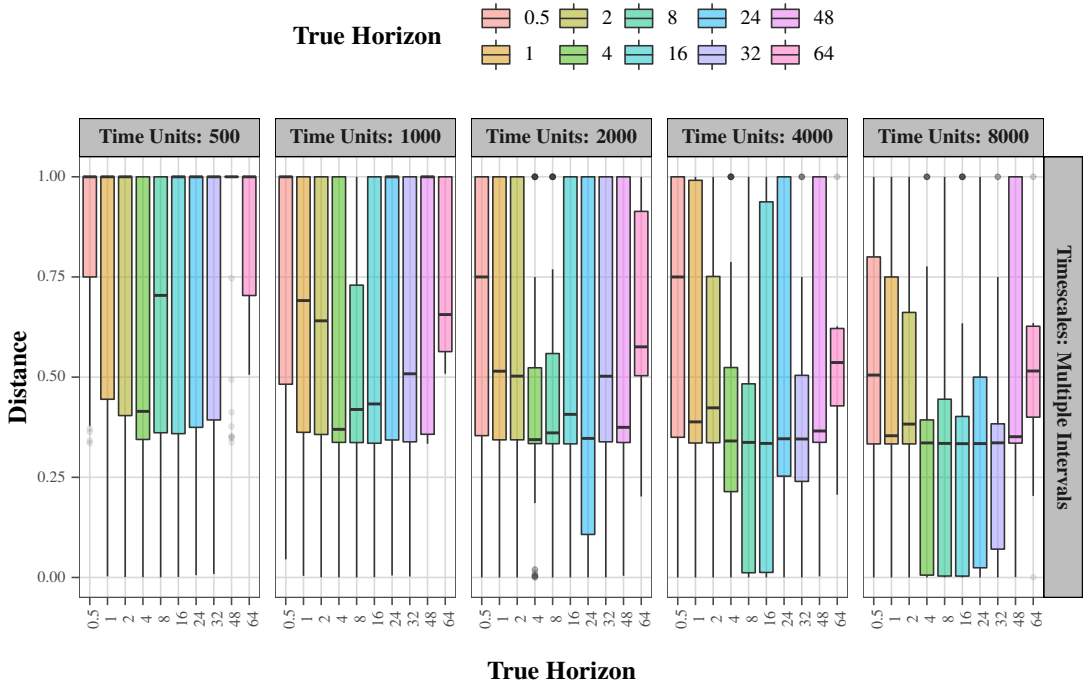}
\caption[Distance per horizon for non-singular timescales]{ Box plots for the distance between edges with multiple interval on their timescales in the data generating TGEMs to the respective edges in the learned TGEMs by the proximal heuristic with respect to the true horizon}
		\label{plot:results_multiple}
\end{figure}

While the distances in Figure \ref{plot:results_single} are reported for edges with a single interval on their timescale, Figure \ref{plot:results_multiple} displays the distances of edges that contained multiple intervals on their timescales. Therefore, they represent a more complex dependency. Moreover, these edges do not only contain the initial horizons, but their temporal range can be extended. In the conducted benchmark, three additional horizons ($32,48,64$) were present in the data generating TGEMs.

Similar to edges with a single interval on their timescale, learning improves with larger data sets and medium temporal ranges tend to be learned better than the extremes.  However, the entire complexity of the dependencies is rarely captured. Rather, the median distance for larger data sets tends to $0.33$ which resembles the case where one interval is \textit{exactly} found but not the other\footnote{Consider a data generating model containing an edge $e$ with  $\mathcal{T} = (0,1], (1,2]$ and a learned model containing an edge $e$ with  $\mathcal{ \widehat{T}} = (0,2]$. In this case, the elementary distance equals $0.33$}.

%% file: figures/average_events.tex
\begin{table}

\caption{\label{tab:avg_events}Summary statistics for event occurrences in the benchmark: Average minimum, median, and maximum events observed  
                per Sampled Time Units}
\centering
\fontsize{8}{10}\selectfont
\begin{tabular}[t]{>{}r||>{}rrr}
\toprule
\multicolumn{1}{c}{ } & \multicolumn{3}{c}{\textbf{Event Occurrences}} \\
\cmidrule(l{3pt}r{3pt}){2-4}
\begingroup\fontsize{8}{10}\selectfont \textbf{Sampled Time Unites}\endgroup & \begingroup\fontsize{8}{10}\selectfont Avg. Min\endgroup & \begingroup\fontsize{8}{10}\selectfont Avg. Median\endgroup & \begingroup\fontsize{8}{10}\selectfont Avg. Max\endgroup\\
\midrule
500 & 10 & 69 & 265\\
1000 & 21 & 143 & 528\\
2000 & 41 & 286 & 1043\\
4000 & 81 & 575 & 2113\\
8000 & 164 & 1157 & 4209\\
\bottomrule
\end{tabular}
\end{table}

%% file: chapters/05_discussion.tex
\section{Conclusion}

The benchmark on synthetic data gained valuable insights for the model class of TGEMs. Generally, the experiments showed that TGEMs can be applied to model a multivariate temporal point process. However, its success strongly depends of the choice of the default horizon. The \textit{proximal heuristic} - an approach that seeks the likelihood-maximizing default horizon -  has been superior to the naive \textit{quantile heuristic} which builds on order statistics. Additionally, temporal dependencies of different length have been reliably detected. With sufficient data the algorithm found short and long temporal dependencies within the same process. However, more complex relations (i.e., with multiple intervals on a timescale) were only partially found - even for large data sets. One explanation might be the parametrization of the benchmark.  If the intervals on the timescale of an edge are large compared to the rate occurrence of the parent node, it is less likely that all configurations are covered in the data sets, thus not providing the necessary variation. In particular, the parent configuration where all intervals are "active" might appear rarely. Hence, the drawn conclusions about the problems to identify more complex relations should be regarded with respect to the settings in the benchmark. Further approaches on synthetic data should therefore consider an even broader set of parameters including a variation of the probability for splits/extends,   or conditional intensity functions with pre-defined behaviour by explicitly assuming amplification or damping rates for given nodes (cf., \citet{Bhattacharjya2018}).

%% file: chapters/appendix_a.tex
\section*{Appendix A}
\label{ch:ApppendigxA}

\subsection*{Random TGEM example}

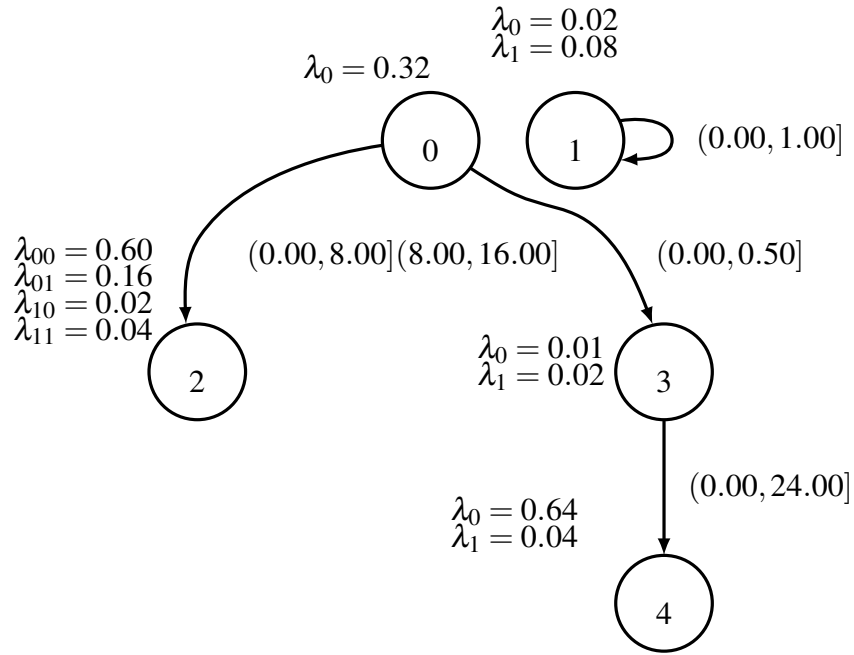
\begin{figure}[htb]
\centering
\input{figures/random_graph.tex}
\caption[Illustration of random TGEM used in benchmark]{Example of a data generating random TGEM used during the benchmark. Edges are denoted with timescales, nodes with their conditional intensity functions}
\label{fig:random_graph}
\end{figure}

\subsection*{Results of the benchmark}

\input{figures/appendix_distance.tex}

\input{figures/appendix_f1.tex}

\begin{figure}[htb]
\centering
\input{figures/benchmark_f1.tex}
\caption[Benchmark results: average $F1$-score per heuristic]{Benchmark results for different default horizon heuristics: Average $F1$-score between the learned model and $100$ data generating TGEMs for various properties (Nodes, Density), and data sets of different sizes}
		\label{plot:results_benchmark_f1}
\end{figure}

%% file: figures/random_graph.tex
\begin{tikzpicture}[>=latex,line join=bevel,transform shape]
  \pgfsetlinewidth{1bp}
\pgfsetcolor{black}
  \draw [->,very thick] (106.92bp,190.14bp) .. controls (86.546bp,186.97bp) and (54.286bp,178.49bp)  .. (39.0bp,156.0bp) .. controls (34.5bp,149.38bp) and (33.053bp,141.08bp)  .. (33.785bp,122.98bp);
  \definecolor{strokecol}{rgb}{0.0,0.0,0.0};
  \pgfsetstrokecolor{strokecol}
  \draw (114.5bp,148.5bp) node {$ (0.00,8.00]   (8.00,16.00]$};
  \draw [->,very thick] (139.89bp,181.51bp) .. controls (143.77bp,178.94bp) and (147.98bp,176.28bp)  .. (152.0bp,174.0bp) .. controls (168.26bp,164.79bp) and (177.02bp,169.44bp)  .. (190.0bp,156.0bp) .. controls (196.35bp,149.42bp) and (201.02bp,140.71bp)  .. (207.74bp,122.86bp);
  \draw (236.5bp,148.5bp) node {$ (0.00,0.50]$};
  \draw [->,very thick] (195.66bp,199.38bp) .. controls (205.62bp,201.02bp) and (215.0bp,198.56bp)  .. (215.0bp,192.0bp) .. controls (215.0bp,187.8bp) and (211.15bp,185.28bp)  .. (195.66bp,184.62bp);
  \draw (251.5bp,192.0bp) node {$ (0.00,1.00]$};
  \draw [->,very thick] (212.0bp,86.974bp) .. controls (212.0bp,75.192bp) and (212.0bp,59.561bp)  .. (212.0bp,36.003bp);
  \draw (251.5bp,61.5bp) node {$ (0.00,24.00]$};
\begin{scope}
  \definecolor{strokecol}{rgb}{0.0,0.0,0.0};
  \pgfsetstrokecolor{strokecol}
  \draw [very thick] (125.0bp,192.0bp) ellipse (18.0bp and 18.0bp);
  \draw (125.0bp,188.3bp) node {$0$};
  \draw (73.0bp,218.8bp) node[right] {$\lambda_0 = 0.32$};
\end{scope}
\begin{scope}
  \definecolor{strokecol}{rgb}{0.0,0.0,0.0};
  \pgfsetstrokecolor{strokecol}
  \draw [very thick] (38.0bp,105.0bp) ellipse (18.0bp and 18.0bp);
  \draw (38.0bp,101.3bp) node {$2$};
  \draw (-35bp,150.6bp) node[right] {$\lambda_{00} = 0.60 $};
  \draw (-35.0bp,140.6bp) node[right] {$\lambda_{01} = 0.16 $};
  \draw (-35.0bp,130.6bp) node[right] {$\lambda_{10} = 0.02 $};
  \draw (-35.0bp,120.6bp) node[right] {$\lambda_{11} = 0.04 $};
  \draw (8.0bp,126.8bp) node[right] {$$};
\end{scope}
\begin{scope}
  \definecolor{strokecol}{rgb}{0.0,0.0,0.0};
  \pgfsetstrokecolor{strokecol}
  \draw [very thick] (212.0bp,105.0bp) ellipse (18.0bp and 18.0bp);
  \draw (212.0bp,101.3bp) node {$3$};
  \draw (174.0bp,112.8bp) node[right] {$ $};
  \draw (138.0bp,113.6bp) node[right] {$\lambda_{0} = 0.01 $};
  \draw (138.0bp,103.6bp) node[right] {$\lambda_{1} = 0.02 $};
  \draw (182.0bp,90.8bp) node[right] {$$};
\end{scope}
\begin{scope}
  \definecolor{strokecol}{rgb}{0.0,0.0,0.0};
  \pgfsetstrokecolor{strokecol}
  \draw [very thick] (179.0bp,192.0bp) ellipse (18.0bp and 18.0bp);
  \draw (179.0bp,188.3bp) node {$1$};
  \draw (141.0bp,235.8bp) node[right] {$ $};
  \draw (143.0bp,236.8bp) node[right] {$\lambda_{0} = 0.02 $};
  \draw (143.0bp,226.8bp) node[right] {$\lambda_{1} = 0.08 $};
  \draw (149.0bp,213.8bp) node[right] {$$};
\end{scope}
\begin{scope}
  \definecolor{strokecol}{rgb}{0.0,0.0,0.0};
  \pgfsetstrokecolor{strokecol}
  \draw [very thick] (212.0bp,18.0bp) ellipse (18.0bp and 18.0bp);
  \draw (212.0bp,14.3bp) node {$4$};
  \draw (128.0bp,52.6bp) node[right] {$\lambda_{0} = 0.64 $};
  \draw (128.0bp,42.6bp) node[right] {$\lambda_{1} = 0.04 $};

\end{scope}
\end{tikzpicture}

%% file: figures/appendix_distance.tex
\begin{table}[H]

\caption[Benchmark results: average distance]{\label{tab:benchmark_dist}Benchmark results on synthetic data sets.
                    Average distance to data generating models per nodes, density, and sampled time units}
\centering
\resizebox{\linewidth}{!}{
\begin{tabular}[t]{rrl>{}r||>{}llllll}
\toprule
\multicolumn{4}{c}{Setting } & \multicolumn{6}{c}{Average Distance (standard deviation) per Horizon Heuristic} \\
\cmidrule(l{3pt}r{3pt}){1-4} \cmidrule(l{3pt}r{3pt}){5-10}
\begingroup\fontsize{10}{12}\selectfont \textbf{Nodes}\endgroup & \begingroup\fontsize{10}{12}\selectfont \textbf{Time Units}\endgroup & \begingroup\fontsize{10}{12}\selectfont \textbf{Density}\endgroup & \begingroup\fontsize{10}{12}\selectfont \textbf{N}\endgroup & \begingroup\fontsize{10}{12}\selectfont \textbf{proximal}\endgroup & \begingroup\fontsize{10}{12}\selectfont \textbf{q =  0.05}\endgroup & \begingroup\fontsize{10}{12}\selectfont \textbf{q =  0.25}\endgroup & \begingroup\fontsize{10}{12}\selectfont \textbf{q =  0.5}\endgroup & \begingroup\fontsize{10}{12}\selectfont \textbf{q =  0.75}\endgroup & \begingroup\fontsize{10}{12}\selectfont \textbf{q =  0.95}\endgroup\\
\midrule
5 & 500 & 0.1 & 100 & {\textbf{1.74}} (1.22) & 2.25 (1.39) & 2.23 (1.34) & 2.25 (1.48) & 2.19 (1.42) & 2.31 (1.45)\\
5 & 1000 & 0.1 & 100 & {\textbf{1.44}} (1.12) & 2.23 (1.37) & 2.17 (1.34) & 2.11 (1.35) & 2.12 (1.5) & 2.23 (1.47)\\
5 & 2000 & 0.1 & 100 & {\textbf{1.26}} (1.03) & 2.17 (1.3) & 2.14 (1.33) & 2.06 (1.32) & 2.04 (1.47) & 2.24 (1.5)\\
5 & 4000 & 0.1 & 100 & {\textbf{1.08}} (0.94) & 2.18 (1.37) & 2.12 (1.42) & 2.04 (1.36) & 2.04 (1.46) & 2.15 (1.5)\\
5 & 8000 & 0.1 & 100 & {\textbf{1.35}} (1.15) & 2.19 (1.37) & 2.12 (1.46) & 2.01 (1.35) & 2.04 (1.48) & 2.17 (1.51)\\
\addlinespace
5 & 500 & 0.2 & 100 & {\textbf{3.55}} (1.84) & 4.34 (1.85) & 4.26 (1.9) & 4.27 (1.97) & 4.34 (2.06) & 4.45 (2.02)\\
5 & 1000 & 0.2 & 100 & {\textbf{2.79}} (1.66) & 4.17 (1.76) & 4.06 (1.77) & 4.19 (1.86) & 4.25 (2.09) & 4.42 (1.95)\\
5 & 2000 & 0.2 & 100 & {\textbf{2.36}} (1.41) & 4.21 (1.82) & 4.03 (1.86) & 4.09 (1.97) & 4.08 (2.01) & 4.19 (1.97)\\
5 & 4000 & 0.2 & 100 & {\textbf{2.05}} (1.27) & 4.1 (1.77) & 3.98 (1.77) & 3.97 (1.98) & 4.04 (2.11) & 4.25 (2.08)\\
5 & 8000 & 0.2 & 100 & {\textbf{2.17}} (1.42) & 4.15 (1.86) & 3.97 (1.81) & 3.92 (1.94) & 3.87 (2.09) & 4.16 (2.15)\\
\addlinespace
10 & 500 & 0.1 & 100 & {\textbf{6.38}} (2.17) & 8.5 (2.38) & 8.5 (2.44) & 8.4 (2.49) & 8.72 (2.62) & 8.91 (2.77)\\
10 & 1000 & 0.1 & 100 & {\textbf{5.22}} (2.01) & 8.42 (2.44) & 8.31 (2.55) & 8.2 (2.47) & 8.52 (2.67) & 8.9 (2.69)\\
10 & 2000 & 0.1 & 100 & {\textbf{4.31}} (2.19) & 8.39 (2.5) & 8.16 (2.58) & 8.01 (2.54) & 8.3 (2.71) & 8.84 (2.7)\\
10 & 4000 & 0.1 & 100 & {\textbf{3.58}} (2.09) & 8.26 (2.41) & 8.11 (2.77) & 7.63 (2.52) & 7.91 (2.59) & 8.64 (2.73)\\
10 & 8000 & 0.1 & 100 & {\textbf{3.68}} (1.95) & 8.39 (2.49) & 8.2 (2.65) & 7.66 (2.6) & 7.78 (2.56) & 8.54 (2.81)\\
\addlinespace
10 & 500 & 0.2 & 100 & {\textbf{10.74}} (2.4) & 14.16 (2.46) & 14.12 (2.52) & 13.99 (2.52) & 14.35 (2.55) & 14.67 (2.37)\\
10 & 1000 & 0.2 & 100 & {\textbf{8.82}} (2.33) & 14.03 (2.3) & 13.58 (2.53) & 13.69 (2.45) & 14.1 (2.9) & 14.46 (2.52)\\
10 & 2000 & 0.2 & 100 & {\textbf{7.29}} (2.55) & 13.91 (2.45) & 13.37 (2.77) & 13.19 (2.75) & 13.71 (2.79) & 14.37 (2.52)\\
10 & 4000 & 0.2 & 100 & {\textbf{6.19}} (2.38) & 13.71 (2.6) & 13.18 (2.89) & 12.89 (2.75) & 13.54 (2.93) & 14.31 (2.81)\\
10 & 8000 & 0.2 & 100 & {\textbf{5.96}} (2.44) & 13.74 (2.71) & 13.04 (3.01) & 12.74 (2.74) & 13.31 (2.91) & 14.02 (2.99)\\
\addlinespace
15 & 500 & 0.1 & 100 & {\textbf{13.71}} (3.32) & 17.33 (3.22) & 17.1 (3.26) & 17.15 (3.28) & 17.39 (3.35) & 17.88 (3.39)\\
15 & 1000 & 0.1 & 100 & {\textbf{10.74}} (2.97) & 16.94 (3.27) & 16.61 (3.36) & 16.61 (3.43) & 17.22 (3.43) & 17.65 (3.27)\\
15 & 2000 & 0.1 & 100 & {\textbf{8.97}} (2.57) & 16.54 (3.13) & 16.23 (3.59) & 16.13 (3.39) & 16.83 (3.62) & 17.46 (3.4)\\
15 & 4000 & 0.1 & 100 & {\textbf{7.56}} (2.63) & 16.35 (3.33) & 15.91 (3.51) & 16.03 (3.66) & 16.39 (3.47) & 17.38 (3.48)\\
15 & 8000 & 0.1 & 100 & {\textbf{7.4}} (2.86) & 16.43 (3.49) & 16.34 (3.69) & 15.78 (3.78) & 16.16 (3.6) & 17.18 (3.69)\\
\addlinespace
15 & 500 & 0.2 & 100 & {\textbf{18.29}} (2.94) & 24.89 (2.29) & 24.64 (2.28) & 24.33 (2.36) & 25.12 (2.84) & 26.14 (2.39)\\
15 & 1000 & 0.2 & 100 & {\textbf{15.07}} (2.35) & 24.37 (2.18) & 24.3 (2.67) & 23.6 (2.62) & 24.5 (2.92) & 25.58 (2.45)\\
15 & 2000 & 0.2 & 100 & {\textbf{12.33}} (2.84) & 24.19 (2.33) & 23.61 (2.52) & 22.69 (2.64) & 23.82 (3.01) & 25.53 (2.52)\\
15 & 4000 & 0.2 & 100 & {\textbf{10.36}} (2.61) & 24.17 (2.28) & 23.43 (2.77) & 22.33 (2.59) & 23.02 (3.21) & 25.05 (2.68)\\
15 & 8000 & 0.2 & 100 & {\textbf{9.14}} (2.66) & 24.28 (2.6) & 23.54 (2.82) & 22.05 (2.46) & 22.49 (3.05) & 24.85 (2.74)\\
\bottomrule
\end{tabular}}
\end{table}

%% file: figures/appendix_f1.tex
\begin{table}[H]

\caption[Benchmark results: average F1-score ]{\label{tab:F1_score}Benchmark results on synthetic data sets.
                    Average F1-score to data generating Models per Nodes, Density, and Sampled Time Units}
\centering
\resizebox{\linewidth}{!}{
\begin{tabular}[t]{rrl>{}r||>{}llllll}
\toprule
\multicolumn{4}{c}{Setting } & \multicolumn{6}{c}{Average F1-Score (standard deviation) per Horizon Heuristic} \\
\cmidrule(l{3pt}r{3pt}){1-4} \cmidrule(l{3pt}r{3pt}){5-10}
\begingroup\fontsize{10}{12}\selectfont \textbf{Nodes}\endgroup & \begingroup\fontsize{10}{12}\selectfont \textbf{Time Units}\endgroup & \begingroup\fontsize{10}{12}\selectfont \textbf{Density}\endgroup & \begingroup\fontsize{10}{12}\selectfont \textbf{N}\endgroup & \begingroup\fontsize{10}{12}\selectfont \textbf{proximal}\endgroup & \begingroup\fontsize{10}{12}\selectfont \textbf{q =  0.05}\endgroup & \begingroup\fontsize{10}{12}\selectfont \textbf{q =  0.25}\endgroup & \begingroup\fontsize{10}{12}\selectfont \textbf{q =  0.5}\endgroup & \begingroup\fontsize{10}{12}\selectfont \textbf{q =  0.75}\endgroup & \begingroup\fontsize{10}{12}\selectfont \textbf{q =  0.95}\endgroup\\
\midrule
5 & 500 & 0.1 & 100 & {\textbf{0.44}} (0.37) & 0.15 (0.26) & 0.22 (0.31) & 0.23 (0.33) & 0.23 (0.33) & 0.14 (0.26)\\
5 & 1000 & 0.1 & 100 & {\textbf{0.57}} (0.37) & 0.17 (0.28) & 0.28 (0.34) & 0.34 (0.36) & 0.32 (0.36) & 0.22 (0.32)\\
5 & 2000 & 0.1 & 100 & {\textbf{0.58}} (0.37) & 0.24 (0.32) & 0.37 (0.34) & 0.4 (0.35) & 0.37 (0.36) & 0.22 (0.32)\\
5 & 4000 & 0.1 & 100 & {\textbf{0.63}} (0.38) & 0.31 (0.34) & 0.45 (0.35) & 0.46 (0.35) & 0.44 (0.36) & 0.32 (0.35)\\
5 & 8000 & 0.1 & 100 & {\textbf{0.61}} (0.37) & 0.38 (0.37) & 0.5 (0.36) & 0.53 (0.35) & 0.47 (0.37) & 0.36 (0.37)\\
\addlinespace
5 & 500 & 0.2 & 100 & {\textbf{0.45}} (0.27) & 0.14 (0.23) & 0.24 (0.27) & 0.26 (0.25) & 0.2 (0.23) & 0.14 (0.22)\\
5 & 1000 & 0.2 & 100 & {\textbf{0.62}} (0.24) & 0.21 (0.25) & 0.36 (0.27) & 0.34 (0.24) & 0.28 (0.26) & 0.15 (0.21)\\
5 & 2000 & 0.2 & 100 & {\textbf{0.68}} (0.24) & 0.27 (0.26) & 0.43 (0.26) & 0.42 (0.27) & 0.39 (0.27) & 0.28 (0.28)\\
5 & 4000 & 0.2 & 100 & {\textbf{0.72}} (0.24) & 0.36 (0.26) & 0.51 (0.26) & 0.5 (0.28) & 0.46 (0.29) & 0.32 (0.29)\\
5 & 8000 & 0.2 & 100 & {\textbf{0.74}} (0.24) & 0.44 (0.26) & 0.59 (0.23) & 0.57 (0.26) & 0.53 (0.28) & 0.4 (0.29)\\
\addlinespace
10 & 500 & 0.1 & 100 & {\textbf{0.58}} (0.15) & 0.22 (0.16) & 0.34 (0.18) & 0.37 (0.17) & 0.29 (0.18) & 0.2 (0.17)\\
10 & 1000 & 0.1 & 100 & {\textbf{0.68}} (0.14) & 0.26 (0.17) & 0.43 (0.19) & 0.44 (0.17) & 0.36 (0.19) & 0.24 (0.18)\\
10 & 2000 & 0.1 & 100 & {\textbf{0.76}} (0.14) & 0.33 (0.17) & 0.51 (0.17) & 0.52 (0.17) & 0.45 (0.16) & 0.28 (0.18)\\
10 & 4000 & 0.1 & 100 & {\textbf{0.81}} (0.12) & 0.43 (0.18) & 0.57 (0.16) & 0.6 (0.14) & 0.54 (0.16) & 0.37 (0.16)\\
10 & 8000 & 0.1 & 100 & {\textbf{0.81}} (0.11) & 0.51 (0.19) & 0.63 (0.14) & 0.65 (0.13) & 0.6 (0.15) & 0.42 (0.16)\\
\addlinespace
10 & 500 & 0.2 & 100 & {\textbf{0.55}} (0.12) & 0.19 (0.13) & 0.3 (0.13) & 0.33 (0.13) & 0.26 (0.12) & 0.18 (0.11)\\
10 & 1000 & 0.2 & 100 & {\textbf{0.66}} (0.1) & 0.24 (0.12) & 0.41 (0.14) & 0.41 (0.12) & 0.33 (0.14) & 0.22 (0.13)\\
10 & 2000 & 0.2 & 100 & {\textbf{0.75}} (0.1) & 0.31 (0.13) & 0.49 (0.14) & 0.49 (0.13) & 0.41 (0.13) & 0.26 (0.13)\\
10 & 4000 & 0.2 & 100 & {\textbf{0.81}} (0.09) & 0.43 (0.13) & 0.57 (0.12) & 0.57 (0.13) & 0.48 (0.14) & 0.33 (0.14)\\
10 & 8000 & 0.2 & 100 & {\textbf{0.83}} (0.09) & 0.51 (0.13) & 0.62 (0.12) & 0.64 (0.1) & 0.55 (0.13) & 0.4 (0.14)\\
\addlinespace
15 & 500 & 0.1 & 100 & {\textbf{0.53}} (0.12) & 0.18 (0.12) & 0.29 (0.13) & 0.3 (0.13) & 0.26 (0.12) & 0.17 (0.11)\\
15 & 1000 & 0.1 & 100 & {\textbf{0.66}} (0.11) & 0.25 (0.13) & 0.4 (0.12) & 0.39 (0.12) & 0.31 (0.13) & 0.21 (0.12)\\
15 & 2000 & 0.1 & 100 & {\textbf{0.74}} (0.09) & 0.33 (0.13) & 0.49 (0.12) & 0.48 (0.12) & 0.4 (0.13) & 0.25 (0.14)\\
15 & 4000 & 0.1 & 100 & {\textbf{0.8}} (0.08) & 0.44 (0.14) & 0.56 (0.11) & 0.55 (0.11) & 0.47 (0.13) & 0.31 (0.14)\\
15 & 8000 & 0.1 & 100 & {\textbf{0.81}} (0.08) & 0.52 (0.11) & 0.61 (0.1) & 0.61 (0.09) & 0.54 (0.12) & 0.38 (0.15)\\
\addlinespace
15 & 500 & 0.2 & 100 & {\textbf{0.57}} (0.1) & 0.2 (0.1) & 0.32 (0.1) & 0.34 (0.1) & 0.26 (0.1) & 0.17 (0.1)\\
15 & 1000 & 0.2 & 100 & {\textbf{0.67}} (0.08) & 0.26 (0.09) & 0.4 (0.09) & 0.42 (0.1) & 0.34 (0.11) & 0.22 (0.1)\\
15 & 2000 & 0.2 & 100 & {\textbf{0.75}} (0.09) & 0.33 (0.09) & 0.5 (0.09) & 0.5 (0.11) & 0.42 (0.1) & 0.25 (0.11)\\
15 & 4000 & 0.2 & 100 & {\textbf{0.81}} (0.07) & 0.42 (0.09) & 0.56 (0.1) & 0.57 (0.1) & 0.5 (0.1) & 0.32 (0.12)\\
15 & 8000 & 0.2 & 100 & {\textbf{0.85}} (0.07) & 0.5 (0.09) & 0.61 (0.1) & 0.63 (0.09) & 0.57 (0.1) & 0.39 (0.11)\\
\bottomrule
\end{tabular}}
\end{table}